\documentclass[journal]{IEEEtran}
\usepackage{cite}

\usepackage[pdftex]{graphicx}
\ifCLASSINFOpdf
\else
\fi
\usepackage{amssymb}

\usepackage{multirow}
\usepackage{array}

\usepackage{svg}

\usepackage{hyperref}       

\usepackage{rotating}
\usepackage{tablefootnote}

\usepackage{pdflscape}

\begin{document}
%
\title{A Survey of End-to-End Driving:\\ Architectures and Training Methods}
%
%
%

\author{Ardi Tampuu,
        Tambet Matiisen,
        Maksym Semikin,
        Dmytro Fishman,
        Naveed Muhammad
\thanks{Manuscript submitted to IEEE Transactions on Neural Networks and Learning Systems special issue on Adaptive Learning and Control for Autonomous Vehicles.}
\thanks{Ardi Tampuu and Tambet Matiisen contributed equally to this work. All authors are with Institute of Computer Science, 
  University of Tartu, Estonia.}
}

\maketitle

\begin{abstract}
Autonomous driving is of great interest to industry and academia alike. The use of machine learning approaches for autonomous driving has long been studied, but mostly in the context of perception. In this paper we take a deeper look on the so called {\em end-to-end} approaches for autonomous driving, where the entire driving pipeline is replaced with a single neural network. We review the learning methods, input and output modalities, network architectures and evaluation schemes in end-to-end driving literature. Interpretability and safety are discussed separately, as they remain challenging for this approach. Beyond providing a comprehensive overview of existing methods, we conclude the review with an architecture that combines the most promising elements of the end-to-end autonomous driving systems.   
\end{abstract}


%
\IEEEpeerreviewmaketitle

%
%
%
%

\section{Introduction}

\IEEEPARstart{A}{utonomy} in the context of robotics is the ability of a robot to operate without human intervention or control \cite{BEKEY_BOOK_2017}. The same definition can be applied to autonomously driving vehicles. The field of autonomous driving has long been investigated by researchers but it has seen a surge of interest in the past decade, both from industry and academia. This surge has been stimulated by DARPA grand and urban challenges of 2004, 2005 and 2007. As such, self-driving cars can be seen as a next potential milestone for the field of artificial intelligence. 

On a broad level, the research on autonomous driving can be divided into two main approaches: (i) modular and (ii) end-to-end. The \emph{modular approach}, also known as the \emph{mediated perception} approach, is widely used by the industry and is nowadays considered the conventional approach. The modular systems stem from architectures that evolved primarily for autonomous mobile robots and that are built of self-contained but inter-connected modules such as perception, localization, planning and control \cite{yurtsever2019survey}. 
As a major advantage, such pipelines are interpretable -- in case of a malfunction or unexpected system behavior, one can identify the module at fault. Nevertheless, building and maintaining such a pipeline is costly and despite many man-years of work, such approaches are still far from complete autonomy.

\begin{figure}
    \centering
    \includegraphics[width=\columnwidth]{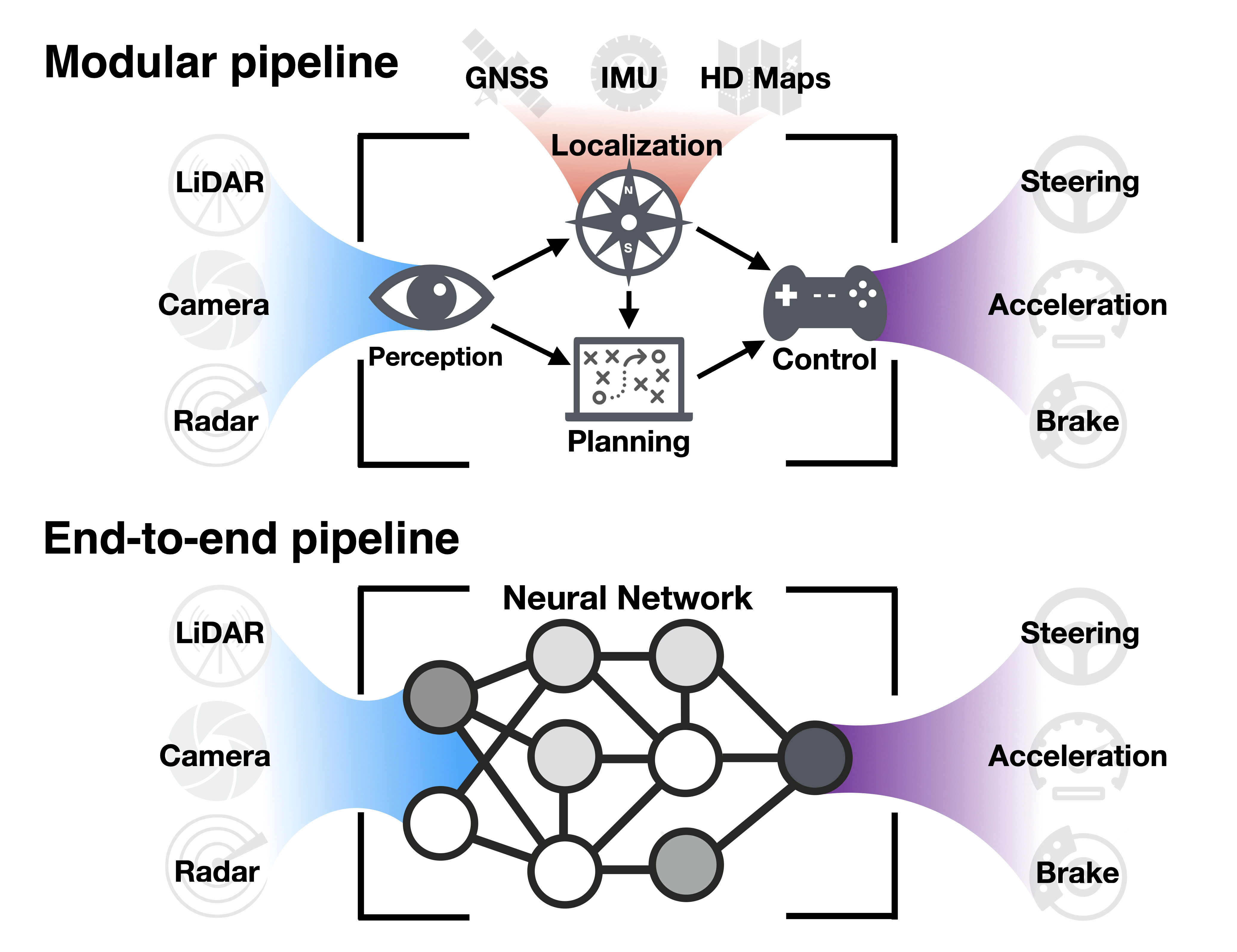}
    \caption{Modular and end-to-end pipelines. The modular pipeline for autonomous driving consists of many interconnected modules, while end-to-end approach treats the entire pipeline as one learnable machine learning task.}
    \label{fig:pipelines}
\end{figure}

\emph{End-to-end driving}, also known in the literature as the \emph{behavior reflex}, is an alternative to the aforementioned modular approach and has become a growing trend in autonomous vehicle research  \cite{pomerleau1989alvinn,muller2006off, bojarski2016end,codevilla2018end,bansal2018chauffeurnet,hecker2018end,kendall2019learning, zeng2019end, chen2019learning}. This approach proposes to directly optimize the entire driving pipeline from processing sensory inputs to generating steering and acceleration commands as a single machine learning task. The driving model is either learned in supervised fashion via \emph{imitation learning} to mimic human drivers \cite{argall2009survey}, or through exploration and improvement of driving policy from scratch via \emph{reinforcement learning} \cite{sutton1998introduction}. Usually, the architecture is simpler than the modular stack with much fewer components (see Figure \ref{fig:pipelines}). While conceptually appealing, this simplicity leads to problems in interpretability -- with few intermediate outputs, it is difficult or even impossible to figure out why the model misbehaves.

Despite a significant amount of research done on, and the promise shown by end-to-end driving approaches, a detailed review of such approaches does not exist. Detailed surveys such as  \cite{yurtsever2019survey} have been carried out in the field of autonomous driving in general, and their scope is very broad. To the best of our knowledge this is the first survey focusing solely on end-to-end approaches to autonomous driving. Hereafter, we define end-to-end driving as a system where a neural network makes the main driving decisions, without constraining what the inputs or outputs of the network are or in how many stages it is trained. 

The survey is organised in terms of different aspects typical to end-to-end driving systems such as learning methods, input and output modalities etc. These aspects are largely independent and if one choice influences the others, it is pointed out in the text. This structure allows the readers to go through the model design choices step-by-step, giving a clear view of existing solutions in each. We begin by comparing end-to-end and modular approaches in the second section, showing the pros and cons of each. In the third section we review the main learning methods or paradigms in end-to-end driving. The fourth section lists potential input modalities used by end-to-end networks and the fifth section considers the possible outputs. The sixth section highlights common evaluation strategies and compares the used metrics. The seventh section reviews different approaches to achieve interpretability, which is commonly seen as a major weakness of neural networks and end-to-end learning. Safety and comfort, discussed in the eighth section, are prerequisites for a real-world use of the self-driving technology. We conclude by discussing the trade-offs provided by different training methods and architectures, and pose a candidate architecture that combines the features used in the reviewed papers. The most relevant of the reviewed works are listed in the Appendix A. The Appendix B contains benchmarking results of models created in the CARLA simulator. The Appendix C lists the potentially useful datasets.

\section{Comparison of modular and end-to-end approaches}

\emph{Modular} approaches involve a fine-grained pipeline of software modules working together to drive the vehicle. The intricate nature of inter-dependencies between such modules is a well-established problem in the broader autonomous-robotics literature (as highlighted in works such as \cite{YANG_WCSE_2017, KREJSA_RECADVM_2010}) and has lead to the development of frameworks such as robot operating system (ROS) \cite{ROS_SOFTWARE}. A detailed description of one such pipeline implementation is presented in \cite{MAHONEY_JFR_2006} which describes the winner of the 2005 DARPA Grand Challenge: a vehicle named Stanley. The Stanley software consisted of around thirty modules, arranged in six layers, namely sensor interface, perception, control, vehicle interface, user interface, and a global service layer. 
The same software architecture was later also employed in 2007 DARPA urban challenge \cite{MONTEMERLO_JFR_2008}. The modular nature of classical pipelines for autonomous driving has also been discussed in \cite{yurtsever2019survey,buehler2009darpa,levinson2011towards,geiger2013vision,paden2016survey,schwarting2018planning, zeng2019end,xiao2019multimodal}.

Modularity enables engineering teams to concentrate on well defined sub-tasks and independently make improvements across the whole stack, keeping the system operational as long as the intermediate outputs are kept functional. However, designing the interconnection of these dozens of modules is an intricate task. Inputs and outputs of each module must be carefully chosen by the engineering teams to accommodate for the final driving task. Similarly, the eventual driving decisions are the result of an ensemble of engineered subsystems that handle different processes based on deterministic rules \cite{buehler2009darpa,levinson2011towards,paden2016survey,yurtsever2019survey}. Clearly defined intermediate representations and deterministic rules make autonomous-driving systems based on modular pipelines behave predictably within their established capabilities, given the strict and known inter-dependencies between different sub-systems. Also, as a major advantage, such pipelines are interpretable. In case of a malfunction or unexpected system behavior, one can track down the initial source of error, e.g. a misdetection \cite{board2018preliminary}. More generally, modularity allows to reliably reason about how the system arrived at specific driving decisions \cite{zeng2019end,xiao2019multimodal,board2018preliminary}.  

However, the modular systems also bear a number of disadvantages. The predefined inputs and outputs of individual sub-systems might not be optimal for the final driving task in different scenarios \cite{zeng2019end}. Different road and traffic conditions might require attending to different pieces of information obtained from the environment. As a result, it is very hard to come up with an exhaustive list of useful information to consider to cover all driving situations. For instance, in terms of perception, commonly the modular stack compresses dynamic objects into 3D bounding boxes. Information not contained in this representation is not retrievable for the subsequent modules. As decision making depends on road and traffic context, the vast variety of driving scenarios and environments makes it incredibly hard to cover all the cases with proper contextualized solutions \cite{dosovitskiy2017carla,zeng2019end}. An example is a ball suddenly rolling to the road. In a residential area, one might expect a child to come retrieve the ball, hence slowing down is the reasonable decision. However, on highways sudden breaking might lead to rear-end collisions and hitting a ball-type object might be less risky than sharp breaking. To deal with both of these situations, first the perception engineers must decide to include the "ball" object-type among the detectable objects in the perception module. Secondly, engineers working on control must label "ball" with different costs depending on the context, resulting in distinct behaviors. There is a long tail of similar, rare but relevant situations - road constructions, accidents ahead, mistakes by other drivers, etc. \cite{dosovitskiy2017carla}. The engineering effort needed to cover all of them with adequate behaviors is immense.  
As a further disadvantage, some of the sub-problems solved in the modular pipeline may be unnecessarily difficult and wasteful \cite{sallab2017deep, zeng2019end}. A perception module trained to detect 3D bounding boxes of all objects on the scene treats all object types and locations equally, just maximizing average precision \cite{zeng2019end}. However, it is clear that in self-driving settings, nearby moving objects are the most crucial to optimize for. Trying to detect all objects leads to longer computation times and trades off precision in relevant objects for good precision overall. Additionally, the uncertainty of detections is often lost when passing detected object boxes to succeeding modules \cite{zeng2019end}.

In \emph{end-to-end} driving the entire pipeline of transforming sensory inputs to driving commands is treated as a single learning task. 
Assuming enough expert driving data for an imitation learning model or sufficient exploration and training for reinforcement learning, the model should learn optimal intermediate representations for the target task. The model is free to attend to any implicit sources of information, as there are no human-defined information bottlenecks. For example, in darkness the presence of an obscured car could be deduced from its headlights reflecting off other objects. Such indirect reasoning is not possible if the driving scene is reduced to object locations (as in modular stack), but is possible if the model can learn by itself what visual patterns to attend to. 

The ability to learn to extract task-specific features and build task-specific representations has lead to a great success of fully neural network based (i.e. end-to-end) solutions in many fields. To begin with, end-to-end approaches are used to solve most computer vision tasks, such as object recognition \cite{he2016deep}, object detection \cite{he2017mask} and semantic segmentation \cite{chen2018encoder}. Neural networks have demonstrated the ability to extract abstract and long-term information from written text and solve tasks such as natural language text generation \cite{radford2019language}, machine translation \cite{vaswani2017attention} and question-answering \cite{lan2019albert}. End-to-end approaches have shown superhuman performance in Atari video games \cite{mnih2015human} and grandmaster level results in highly competitive multiplayer games such as StarCraft \cite{vinyals2019grandmaster} and Dota 2 \cite{berner2019dota}. End-to-end neural networks have been also the crucial component in conquering board games such as Go \cite{silver2016mastering} and Chess \cite{silver2017mastering}. Many of these solved tasks are of similar complexity to driving a car, in terms of the number of possible states and the long term planning needed. Consider that driving is a task that a large proportion of people successfully perform even when tired or distracted. Oftentimes a person can later recollect very little about the route, suggesting the task is a behavior reflex task that demands little conscious attention. Such tasks that do not need higher-level reasoning should be more solvable for end-to-end approaches. It is, therefore, reasonable to believe that in the near future an end-to-end approach is also capable to autonomously control a vehicle.

The use of end-to-end optimization raises an issue of interpretability. With no intermediate outputs, it is much harder to trace the initial cause of an error as well as to explain why the model arrived at specific driving decisions \cite{xiao2019multimodal,bojarski2017explaining, kim2017interpretable,zeng2019end}. However, solutions exist to increase the interpretability of end-to-end driving models (discussed later in Section \ref{sec:interpretability}). 

While powerful and successful, it is well known that neural networks are susceptible to adversarial attacks \cite{szegedy2013intriguing}. By making small, but carefully chosen changes to the inputs, the models can be fooled and tricked into making errors. This leads to serious safety concerns when applying neural networks to high-risk domains such as autonomous driving (addressed in Section \ref{sec:safety}).

Despite having received less interest and investment over the years, the end-to-end approach to autonomous driving has recently shown promising results. While interpretability remains a challenge, the ease of training and the simplicity of the end-to-end models are appealing. The success of neural networks in other domains also supports continued interest in the field.  

\section{Learning methods}
\label{sec:learning}

Below we describe the common learning methods in end-to-end driving.

\subsection{Imitation learning}

\emph{Imitation learning} (IL) or \emph{behavior cloning} is the dominant paradigm used in end-to-end driving. Imitation learning is a supervised learning approach in which a model is trained to mimic expert behavior \cite{argall2009survey}. In the case of autonomous driving, the expert is a human driver and the mimicked behavior is the driving commands, e.g. steering, acceleration and braking. The model is optimized to produce the same driving actions as a human, based on the sensory input recorded while the human was driving. The simplicity of collecting large amounts of human driving data makes the IL approach work quite well for simple tasks such as lane following \cite{pomerleau1989alvinn, bojarski2016end}. However, more complicated and rarely occurring traffic scenarios remain challenging for this approach \cite{codevilla2019exploring}. 

The first use of imitation learning for end-to-end control was the seminal ALVINN model by Dean Pomerleau \cite{pomerleau1989alvinn}. In that work, a shallow fully-connected neural network learned to predict steering wheel angle from camera and radar images. The NAVLAB car steered by ALVINN was able to perform lane following on public roads. Obstacle avoidance was first achieved by Muller \emph{et al.} \cite{muller2006off} with a small robot car, DAVE, navigating in a cluttered backyard. DAVE used two cameras enabling the network to extract distance information. More recently, NVIDIA \cite{bojarski2016end} brought the end-to-end driving paradigm to the forefront by training a large-scale convolutional neural network to steer a commercial vehicle in a range of driving conditions including both highways and smaller residential roads. 

\subsubsection*{Distribution shift problem}

An imitation learning model learns to mimic the experts’s response to traffic situations that the expert has caused. In contrast, when the car is driven by the model, the model’s own outputs influence the car’s observations in the next time step. Hence, the model needs to respond to situations its own driving leads to. If the driving decisions lead to unseen situations, the model might no longer know how to behave. 

Self-driving leading the car to states unseen during training is called the \emph{distribution shift} problem \cite{ross2011reduction,codevilla2019exploring} - the actual observations when driving differ from the expert driving presented during training. For example, if the expert always drove near the center of the road, the model has never seen how to recover when deviating towards the side of the road.

Potential solutions to the distribution shift problem are: data augmentation, data diversification and on-policy learning. All these methods diversify the training data in some way - either by collecting or generating additional data. Diversity of training data is crucial for generalization.

\subsubsection{Data augmentation}
\begin{figure}
    \centering
    \includegraphics[width=\columnwidth]{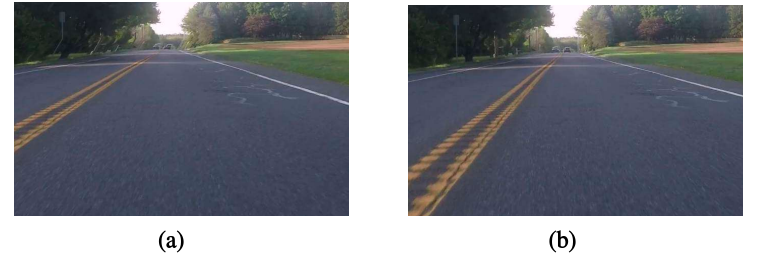}
    \caption{ (a) An original and (b) a synthesized image from \cite{bojarski2016end}. The synthesized image looks as if the car has drifted towards the center of the road. }
    \label{fig:viewpoint}
\end{figure}

Collecting a large and diverse enough dataset can be challenging. Instead, one can generate additional, artificial data via \emph{data augmentation} \cite{shorten2019survey}. Blurring, cropping, changing image brightness and adding noise to the image are standard methods and have also been applied in self-driving context \cite{sobh2018end}. Furthermore, original camera images can be shifted and rotated as if the car had deviated from the center of the lane (see Figure \ref{fig:viewpoint}) \cite{pomerleau1989alvinn, bojarski2016end, codevilla2018end}. The artificial images need to be associated with target driving commands to recover from such deviations. Such artificial deviations have been sufficient to avoid accumulation of errors in lane-keeping. Additionally, one can place two additional cameras pointing forward-left and forward-right and associate the images with commands to turn right and left respectively \cite{bojarski2016end,muller2018driving}.

\subsubsection{Data diversification}
\begin{figure}[t]
    \centering
    \includegraphics[width=\columnwidth]{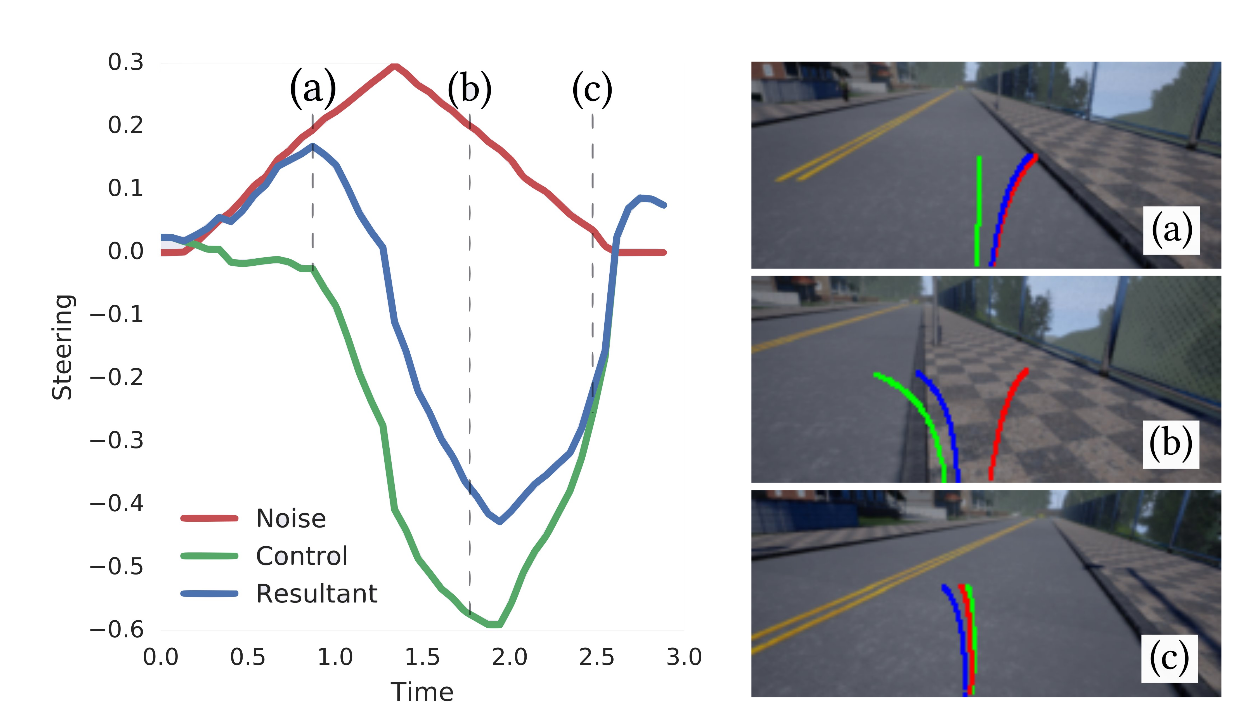}
    \caption{ Data diversification in \cite{codevilla2018end}. Noise is injected during data collection. \textbf{Left:} steering control in [rad/s]. The steering signal provided to the car (blue) is the sum of the driver’s (green) control and the noise (red). \textbf{Right:} driver’s point of view at three points in time (the trajectories are added for visualization).  (a) the  noise  starts to produce  a  drift  to  the  right. (b) This triggers a human reaction, sharp turn to the left. (c) Finally, the car recovers from the disturbance. The driver-provided signal is used for training.}
    \label{fig:noise}
\end{figure}

In addition to data augmentation, it is possible to \emph{diversify the training data} during collection \cite{codevilla2018end,dosovitskiy2017carla,zhou2019does, codevilla2019exploring,muller2018driving, sobh2018end}. Noise can be added to the expert's driving commands at recording time. The noise forces the car off trajectory, forcing the expert to react to the disturbance and by doing so provide examples of how to deal with deviations.  While noise uncorrelated across timesteps might be sufficient \cite{sobh2018end}, often temporally correlated noise (see Figure \ref{fig:noise}) is added \cite{codevilla2018end,dosovitskiy2017carla,zhou2019does, codevilla2019exploring,muller2018driving}. While useful in simulation, such diversification technique might be too dangerous to apply in the real world. 

\begin{figure}[t]
    \centering
    \includegraphics[width=0.9\columnwidth]{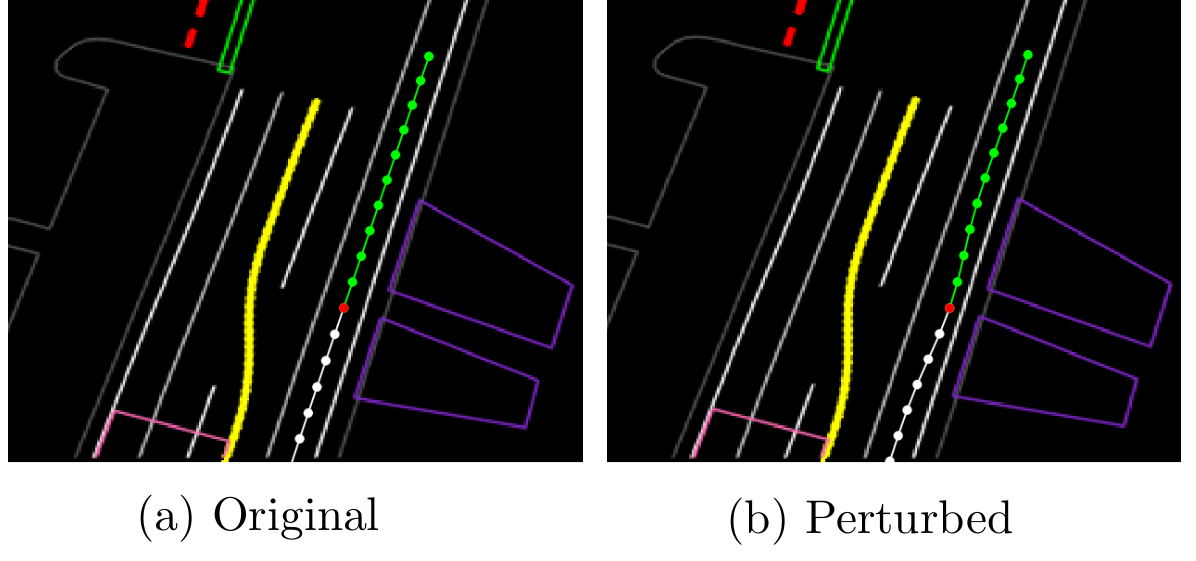}
    \caption {{Trajectory Perturbation as illustrated in \cite{bansal2018chauffeurnet}, where dots connected by a line form a trajectory. (a) Original unperturbed trajectory. The vehicle drives at the center of a lane. (b) A perturbed trajectory, obtained by shifting an agent location (red dot) away from the center of lane and then fitting a line that brings the agent back to the center (first few green dots after the red).}}
    \label{fig:perturb}
\end{figure}

ChauffeurNet \cite{bansal2018chauffeurnet} predicts future waypoints from top-down semantic images instead of camera images. The system uses synthetic trajectory perturbations (including collisions, going off the road) during training for more robust policy (see Figure \ref{fig:perturb}). Due to the semantic nature of the inputs, perturbations are much easier to implement than with camera inputs.

\subsubsection{On-policy learning}

To overcome the model not being able to correct course when drifting away from the center of the lane, DAgger \cite{ross2011reduction} proposes to alternate between model and expert while collecting driving data. In principle, the expert provides examples how to solve situations the model-driving leads to. Such systems learn to recover from errors and handle situations that would never happen during human driving. Recovery annotations can be recorded offline \cite{ross2011reduction} or online by having an expert override the control when the autonomously controlled car makes mistakes \cite{chowdhuri2019multinet}.

However, keeping an expert in the loop is expensive, the amount of possible ways to fail is unlimited. SafeDAgger \cite{zhang2016query} reduced human involvement by a module deciding if the self-driving model needs expert’s help at any given moment. The need for human labels can be further reduced by using automatic methods to recover from bad situations. For example, OIL (observational imitation learning) \cite{li2018oil} uses both a set of localization based PID (proportional-integral-derivative) controllers and a human as experts to train IL policies. Taking a step further, one can completely remove the need for human experts in DAgger by applying a conventional localization and controller stack to automatically annotate the states encountered \cite{pan2018agile}.

When using an algorithm as the expert that provides the true labels \cite{pan2018agile,chen2019learning}, one can do imitation learning on-policy -- the learning agent controls the car during data collection while continuously getting supervision from the expert algorithm. Furthermore, the learned policy can rely on a different (e.g. more cost-effective) set of sensors than the expert algorithm. With these insights, Chen \emph{et al.} \cite{chen2019learning} first trained a \emph{privileged agent} with access to ground-truth road map to imitate expert autopilot (off-policy). They then used the trained \emph{privileged agent} as the expert to train on-policy a \emph{sensorimotor agent} with only visual input. Learning on-policy from the privileged agent instead of learning off-policy from the original IL labels resulted in a drastically improved vision-only driving model. Additionally, for a model with multiple branches corresponding to different navigational commands (more details in section \ref{sec:inputs}), the target values for each possible command can be queried from the privileged agent and all branches receive supervision at all times.

\subsubsection*{Dataset balancing}

Biases in the training dataset can reduce machine learning models' test-time performance and generalization ability \cite{torralba2011unbiased,binns2017fairness}. Imitation learning for self-driving is particularly susceptible to this problem, because 1) the datasets are dominated by common behaviors such as driving straight with stable speed \cite{codevilla2019exploring} and 2) the inputs are high-dimensional and there is plenty of spurious correlations that can lead to causal confusion \cite{de2019causal}. At the same time the self-driving task contains many very rarely occurring and difficult to solve situations. When optimizing for average imitation accuracy, the model might trade-off performance on rare hard-to-solve cases for imitating precisely the easy common cases. As a result, training on more data can actually lead to a decrease of generalization ability, as experienced in \cite{codevilla2019exploring}.

Balancing the dataset according to steering angle can help remedy the inherent biases. Balancing can be achieved via upsampling the rarely occurring angles, downsampling the common ones or by weighting the samples \cite{chawla2009data}. Hawke \emph{et al.} \cite{hawke2019urban} divide the steering angle space into bins and assign the sample weights to be equal to the bin width divided by the number of points in the bin. This leads to the few samples in sparse but wide bins having increased influence compared to samples from a densely populated narrow bin. Furthermore, within each bin they similarly bin and weight samples according to speed. The balancing of training data according to the two output dimensions (speed and steering) proved so effective that the authors claim further data augmentation or synthesis was not needed \cite{hawke2019urban}.

Instead of balancing according to output dimensions, one can also balance the dataset according to data point difficulty. One such approach is to change the sampling frequency of data points according to the prediction error the model makes on them. This has been applied in imitation learning models \cite{chen2019learning}. Alternatively, instead of re-sampling by error, one could also use weighting by error (weight the hard-to-predict data points higher).

The need for dataset balancing might also arise in case of incorporating navigational commands such as ``turn left'', ``turn right'' and ``go straight'' (see Section \ref{sec:navinputs}), the latter being by far the most common in recorded datasets. Balancing the mini-batches to include the same amount of each command is a proposed solution \cite{dosovitskiy2017carla}. 

\subsubsection*{Training instability}

Another challenge of using neural networks as driving models is that the training process is unstable and is not guaranteed to converge to the same minima each time \cite{neal2018modern}. In fact, with different network initialization or with different ordering of training samples into batches, the trained self-driving model might obtain qualitatively different driving behaviors \cite{codevilla2019exploring}. Minor differences in model outputs are amplified at test time when the model's actions define its future inputs, resulting in completely different behaviors. As all other neural networks, end-to-end models are sensitive to dataset biases and may overfit \cite{codevilla2019exploring}.

\subsection{Reinforcement learning}
\emph{Reinforcement learning} (RL) is a machine learning paradigm where a system learns to maximize the rewards given to it by acting in an environment \cite{sutton1998introduction}. While IL often suffers from insufficient exposure to diverse driving situations during training, RL is more immune to this problem. The learning happens online, so sufficient exploration during the training phase leads to encountering and learning to deal with the relevant situations. However, RL is known to be less data efficient than IL and also challenging to use in the real world. 

The inputs for a RL driving policy can be the same as for IL models. When using policy gradient methods, the outputs can also be the same as in IL, no change in network architecture is needed. In RL the learning signal originates from rewards, which need to be computed and recorded at each time step. There is no need to collect expert driving recordings, i.e. human labeled data. 

In an early work, Reidmiller \emph{et al.} \cite{riedmiller2007learning} applied reinforcement learning to learn steering on a real car based on five variables describing the state of the car. More recently, Koutnik \emph{et al.} \cite{koutnik2014evolving} learned to drive based on image inputs in the TORCS game environment \cite{wymann2000torcs} via neuroevolution. RL was also used to learn the difficult double-merge scenario with multiple agents \cite{shalev2016safe} each trained by a policy gradient algorithm \cite{williams1992simple, sutton2000policy}. Another popular RL approach, deep Q-learning, has also been used to train vision-based driving in simulation \cite{wolf2017learning}. Deep deterministic policy gradients (DDPG, \cite{lillicrap2015continuous}) has also been applied to self driving \cite{liang2018cirl,kendall2019learning}, as well as proximal policy optimization (PPO \cite{schulman2017proximal}) in \cite{osinski2019simulation}. 

Policies can be first trained with IL and then fine-tuned with RL methods \cite{liang2018cirl}. In other words, one initializes the RL policy with an IL-trained model. This approach reduces the long training time of RL approaches and, as the RL-based fine-tuning happens online, also helps overcome the problem of IL models learning off-policy. 

\subsubsection{Rewards}

While in imitation learning the desired behavior is defined by the ground truth actions, in reinforcement learning the model simply aims to maximize the rewards. Therefore, the choice of positively and negatively rewarded events and actions influences the eventual learned behavior very directly. The simpler the rewarding scheme, the easier it is to understand why certain (mis)behaviors emerge. However, the more complicated rewarding scheme is used, the more explicitly we can define what is desirable and what is not. 

The most common is to reward the movement speed (towards the goal; along the road) \cite{pan2017virtual,dosovitskiy2017carla, liang2018cirl,kendall2019learning}. Another option is to penalize not being near the center of the track \cite{riedmiller2007learning,pan2017virtual} or a reference trajectory \cite{osinski2019simulation}. While crashes shorten the distance covered without disengagement and are implicitly avoided when maximizing discounted future reward (e.g. in \cite{kendall2019learning}), one might also just explicitly punish crashes by assigning a large negative reward, as was done in \cite{michels2005high, liang2018cirl,pan2017virtual,dosovitskiy2017carla,shalev2016safe}. Additionally, punishments for overlap with sidewalks and the opposite lane \cite{liang2018cirl, dosovitskiy2017carla} and for abnormal steering angles  \cite{liang2018cirl} have been used. Multiple authors find that combining more than one of these rewards is beneficial \cite{dosovitskiy2017carla,pan2017virtual,liang2018cirl}. 

\subsubsection{Learning in the real world}
Crucial challenge in training reinforcement learning policies lies in providing the necessary exploration for driving policies without incurring damage on the vehicle or other objects. \emph{Learning in the real world} is still possible by adding a safety measure that takes over control if the RL policy deviates from the road. Riedmiller \emph{et al.} \cite{riedmiller2007learning} used an analytically derived steering controller for taking control of the car in case the RL policy deviated too much from the center of track. As another option, a safety-driver can be used if using real cars \cite{kendall2019learning}.

\subsubsection{Learning in simulation}
Alternatively, it is common to train and test the model \emph{within a simulated environment}. Recently, the CARLA simulator \cite{dosovitskiy2017carla} and GTA V computer game \cite{zhou2019does} have been used for training IL and RL models. These engines simulate an urban driving environment with cars, pedestrians and other objects. Beyond creating the CARLA simulator, Dosovitsky \emph{et al.} \cite{dosovitskiy2017carla} also trained and evaluated an asynchronous advantage actor-critic (A3C) deep reinforcement learning method. Despite training the RL model more extensively, the imitation learning models and a classical modular pipeline outperformed the RL-based method at evaluation \cite{dosovitskiy2017carla}.

\subsection{Transfer from simulation to real world}
To avoid costly crashes and endangering humans by using real cars, one can train a model in simulation and apply it in the real world \cite{michels2005high}. However, as the inputs from simulations and from the real world are somewhat different, the generalization might suffer if no means are taken to minimize the difference in input distributions or to adapt the model. The problem of adapting a model to new, different data, can be approached via supervised \cite{sharif2014cnn,yosinski2014transferable} or unsupervised learning \cite{ganin2014unsupervised,tzeng2017adversarial}. 

\emph{Fine-tuning} \cite{sharif2014cnn,yosinski2014transferable} is a common supervised approach for adapting models to new data. A model trained in simulation can be re-trained using some real world data. The model can thus adapt itself to the new input distribution. Labeled examples from the real world are needed for this adapting. This approach, however, has not commonly been used in end-to-end driving.

Using an unsupervised approach, one can instead adapt the incoming data and keep the driving model fixed. This approach requires only unlabeled data from the real world, not labeled data. Using conditional generative adversarial networks (cGANs) \cite{goodfellow2014generative,mirza2014conditional}, real-looking images can be generated based on images from a simulator \cite{pan2017virtual}. An end-to-end model can then be trained on the synthetic made-to-look-real images. The resulting driving model can be deployed in the real world or just evaluated by comparing to real-world recordings \cite{pan2017virtual}.

Conversely, real images can be transformed into simulation-like images via a cGAN \cite{zhang2019vr}. Generating real-looking images is challenging, generating simulation-like images can be easier. With this approach, driving models can be trained in simulation and used on real data by adapting the inputs. When evaluated in the real world, models trained on simulated-to-real transformed images do not need an adaptation module, whereas models trained on real-to-simulated transformed images need the domain-adaptation module. This means the former is computationally more efficient on real world data. However, the latter is more efficient to train in simulation.

\begin{figure}
    \centering
    \includegraphics[width=\columnwidth]{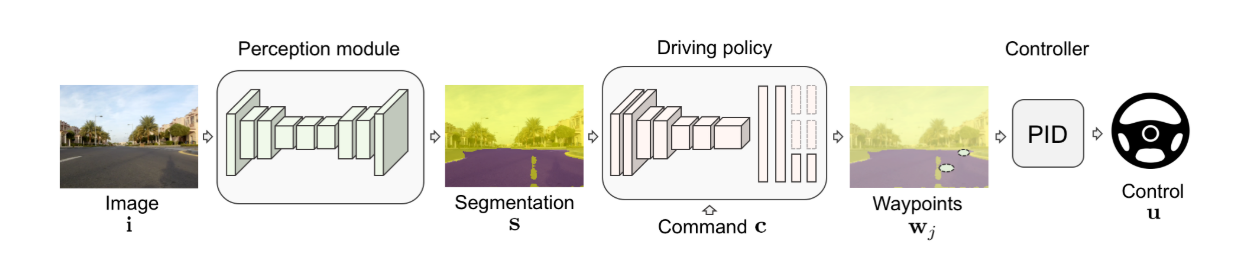}
    \caption{In Müller \emph{et al.} \cite{muller2018driving} the simulated environment image is turned into segmentation map which is in turn used by the driving policy. Both simulated and real images can be segmented. The driving policy does not care where the segmentations originated from and can drive in both environments.}
    \label{fig:sim2real_muller}
\end{figure}

Both real and simulated images can be mapped to a common representation that is informative, but sufficiently abstract to remove unnecessary low-level details. When training a driving model using this representation as input, the behavior does not depend any more on where the inputs originate from. Müller \emph{et al.} \cite{muller2018driving} proposed to extract the semantic segmentation of the scene from both the real and the simulated images and use it as the input for the driving policy. This approach allowed to successfully transfer a policy learned in CARLA simulator to a 1/5 sized real-world truck. The processing pipeline is demonstrated in Figure \ref{fig:sim2real_muller}.

{}

\begin{figure}
    \centering
    \includegraphics[width=0.9\columnwidth]{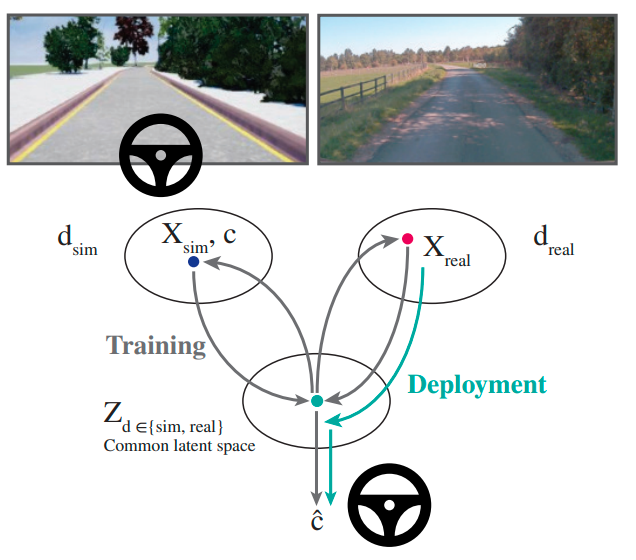}
    \caption{Bewley \emph{et al.} \cite{bewley2019learning} use two variational autoencoders to learn a common latent space between simulated and real images. An image from either domain can be translated to the common latent space. The latent representation is used to generate steering command, but can also be used to generate an image in either of the original domains. This allows to learn a common representation via cyclic losses. For example cyclic reconstruction loss compares simulated images with images obtained by 1) encoding this simulated image to latent space, then 2) generating a "real image" from the resulting vector, then 3) mapping the generated image again to latent space and 4) generating an artificial simulated image from the new latent vector. }
    \label{fig:sim2real_bewley}
\end{figure}

Instead of specifying an arbitrary intermediate representation (e.g. semantic segmentation), end-to-end optimization can be made to learn the most useful latent representation from the data \cite{bewley2019learning}. To do that, domain transfer modules learn to map images from simulation and real world into a shared lower-dimensional latent space (and back) via two variational autoencoders \cite{kingma2013auto, diederik2014auto,liu2017unsupervised}, one per domain (see Figure \ref{fig:sim2real_bewley}). The unsupervised transfer modules do not need corresponding pairs of simulation and real-world images for training. The driving policy is trained with IL in simulation. The same latent-to-steering policy is later applied to the real-world by using the real-image-to-latent encoder. The described method \cite{bewley2019learning} was tested on a real vehicle and outperformed other domain adaptation methods, including  simulation-to-real \cite{muller2018driving} and real-to-simulation \cite{zhang2019vr} transformation.

Simulations can be useful also to test and compare architectures, combinations of inputs, fusion methods etc. Hopefully, what works well on simulated data also works on real inputs. Kendall \emph{et al.} \cite{kendall2019learning} used simulations in Unreal engine to select architecture, action space and learning hyper-parameters before training an RL model in the real world.

\section{Input modalities}
\label{sec:inputs}
Visual inputs combined with intrinsic knowledge about the world and the desired route are usually sufficient for a human driver to safely navigate from point A to point B. External guidance from route planers is often also visual, though voice commands are useful to keep attention on the road. In self-driving, however, the range of possible input modalities is wider. Different inputs can complement each other and help improve generalization and accuracy \cite{xiao2019multimodal}. Hence, while vision could be sufficient to drive, in end-to-end driving often multiple input modalities are used simultaneously. Most commonly used modalities are described as follows. 

\subsection{Camera vision}

Monocular camera image is the most natural input modality for end-to-end driving and was already used in \cite{pomerleau1989alvinn}. Indeed, humans can also drive with vision in only one eye \cite{wood2002aging,racette2005impact}, hence stereo-vision is not a pre-requisite of driving. Many models have managed to achieve good performance with monocular vision-only models \cite{ kendall2019learning, bojarski2016end, codevilla2018end, osinski2019simulation}. Nevertheless, other authors have found it useful to use stereo cameras, allowing convolutional neural networks (CNNs) to learn to implicitly extract depth information \cite{muller2006off, chowdhuri2019multinet}. 

To model temporal aspects, the model needs to consider the combination of multiple past frames \cite{sauer2018conditional,hecker2018end,chowdhuri2019multinet,zhou2019does,hawke2019urban}. 

Surround-view cameras are necessary for lane changes and for giving way on intersections. They achieve the function of rear-view mirrors used by human drivers \cite{hecker2018end}. Surround-view led to an improved imitation accuracy on turns and intersections, but no improvement was seen on highways \cite{hecker2018end}. 

While mostly the self-driving datasets are sufficiently large to train end-to-end networks with millions of parameters from scratch, it is possible to use pre-trained object-detection networks as out-of-the-box feature extractors \cite{chen2019learning,hecker2018end, xu2017end}. Different versions of ResNet \cite{he2016deep} trained on ImageNet dataset are commonly used. The pre-trained layers can be fine-tuned for the current task \cite{xu2017end,hecker2019learning}. 

\subsection{Semantic representations}
Instead or in addition to the RGB images one can also provide the model with extracted representations of the visual scene such as semantic segmentation, depth map, surface normals, optical flow and albedo \cite{zhou2019does,muller2018driving, sobh2018end, hawke2019urban}. In simulation, these images can be obtained with perfect precision. Alternatively (and in the real world) these inputs can be generated from the original images by using other, specialized networks. The specialized networks, e.g. for semantic segmentation, can be pre-trained on existing datasets and do not need to be trained simultaneously with the driving model. While the original RGB image contains all the information present in the predicted images, explicitly extracting such pre-defined representations and using them as (additional) inputs has been shown to improve model robustness \cite{zhou2019does,sobh2018end}. 

\subsection{Vehicle state}
Multiple authors have additionally provided their models with high-level measurements about the state of the vehicle, such as current speed and acceleration \cite{codevilla2018end,liang2018cirl,caltagirone2017lidar,xu2017end,osinski2019simulation}.

Current speed is particularly useful when the model does not consider multiple consecutive frames. However, in imitation learning settings, if the model receives the current speed as input and predicts the speed for the next timestep as one of the outputs, this can lead to the \emph{inertia problem} \cite{codevilla2019exploring}. As in the vast majority of samples the current and next timestep speeds are highly correlated, the model learns to base its speed prediction exclusively on current speed. This leads to the model being reluctant to change its speed, for example to start moving again after stopping behind another car or a at traffic light \cite{codevilla2019exploring}. This problem can be remedied by helping the model learn speed-related internal representations from images via an additional speed output (not used for controlling the car), but has not been solved completely. 

\subsection{Navigational inputs}
\label{sec:navinputs}

Being able to follow lane and avoid obstacles are crucial parts of any self-driving model. However, a self-driving car becomes actually useful once we are able to decide where to drive. The indispensable feature of choosing where the car will take you has been studied also in end-to-end models. Multiple approaches have been applied.

\subsubsection{Navigational commands}
Navigating from point A to point B can be achieved by providing an additional navigational command assuming values like “go left”, “go right”, “go straight” and "follow the road" \cite{codevilla2018end,sauer2018conditional,liang2018cirl,xiao2019multimodal,hawke2019urban,chen2019learning,caltagirone2017lidar}. These commands are generated by a high-level route planner. Notice that similar commands are sufficient for human drivers and are often provided by navigation tools. The prevailing method to incorporate these commands is to have separate output heads (or branches of more than one layer) for each command and just switch between them depending on the received command. Codevilla \emph{et al.} \cite{codevilla2018end} demonstrated that such approach works better than just inputting the navigational command to the network as an additional categorical variable. However, such an approach does not scale well with increasing number of different commands. Using navigation commands as additional inputs to intermediate layers \cite{hawke2019urban,chowdhuri2019multinet} is an alternative. Similarly to switching branches, this allows the conditional input to be closer to the model output, influencing more directly decision-making, not scene understanding. 

As a generalization of navigational commands, driving models can be requested to drive in a certain manner. Similarly to inserting a command to turn left or right, a model can be ordered to keep to the right of the lane, to the center of the lane or follow another vehicle \cite{chowdhuri2019multinet} via additional categorical inputs. In a similar manner, switching between slow, fast, aggressive or cautious modes can be envisaged. Learning personalized driving styles was investigated in \cite{kuderer2015learning}.

\subsubsection{Route planner}

“Left”, “right” and “straight” navigational inputs are momentary and do not allow long-term path planning. A more informative representation of the desired route can be inserted to the model in the form of visual route representations. Intuitively, navigation app screen images can be used as input \cite{hecker2018end}, similarly to what a human driver would see when using the planner. Adding a route planner screen as a secondary input to end-to-end models yields higher imitation accuracy compared to models without route input. Using a raw list of desired future global positioning system (GPS) coordinates as input was not as effective as using a route planner screen \cite{hecker2018end}. 

\begin{figure}
    \centering
    \includegraphics[height=2.4cm]{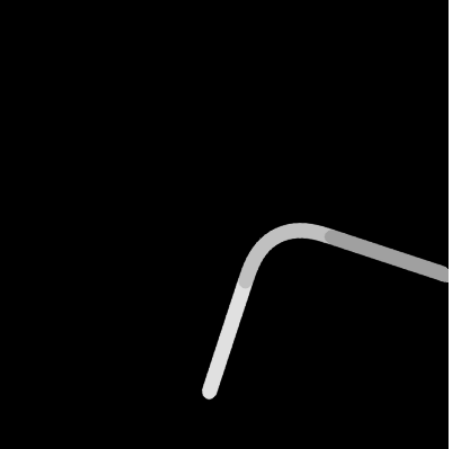}
    \includegraphics[height=2.4cm]{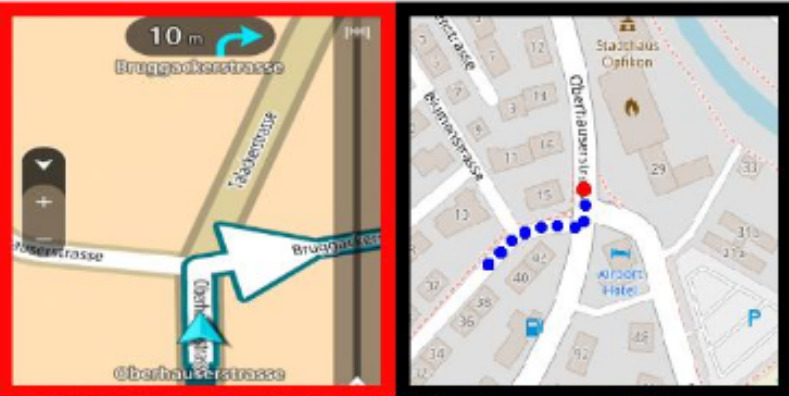}
    \caption{\textbf{Left:} Route input in ChauffeurNet \cite{bansal2018chauffeurnet} is a binary map. \textbf{Middle and right:} Route input in Hecker \emph{et al.} \cite{hecker2018end} is either a screen image from TomTom (middle) or a list of GPS coordinates (right).}
    \label{fig:route}
\end{figure}

ChauffeurNet \cite{bansal2018chauffeurnet} operates on top-down high-definition (HD) maps and provides also the desired route as a binary top-down image (Figure \ref{fig:route}), i.e the same frame of reference as other inputs. 
{}

\subsubsection{Textual commands}
Using text as additional input for driving policies has also been explored. The extra advice can either be goal-oriented such as “Drive slowly” or descriptive such as “There is a pedestrian”. Textual advice can help the policy to better predict expert trajectories compared to camera-only approaches. \cite{kim2019grounding}

\subsection{LiDAR}

Another notable source of inputs in self-driving is LiDAR. LiDAR point clouds are insensitive to illumination conditions and can provide good distance estimations. The output from LiDAR is a sparse cloud of unordered points, which needs to be processed to extract useful features. It is very common to preprocess these points into binary 3D occupancy grids and input them to CNNs \cite{zeng2019end,casas2018intentnet,caltagirone2017lidar}. However, working with the 3D occupancy grid (e.g. via 3D convolution) can be costly \cite{zhou2018voxelnet}, because the number of voxels increases rapidly with spatial precision. There are however methods that allow to reduce this processing time, for example by applying convolutions sparsely, only on locations associated with input points \cite{yan2018second}.
Instead of using 3D occupancy grids, PointNet \cite{qi2017pointnet, qi2017pointnet++} allows to convert the raw point cloud directly to usable features, as used in \cite{chen2018lidar}.



A further option is to project the point cloud to a 2D grid in bird's-eye view (BEV) \cite{chen2017multi,simon2018complex}. 
Other sources of information can also be mapped to the BEV reference frame \cite{casas2018intentnet,bansal2018chauffeurnet}, which can make fusion of raw inputs or extracted CNN feature maps more convenient.

LiDAR point cloud can be mapped to a 2D image also via polar grid mapping (PGM) \cite{sobh2018end}. PGM outputs images where each column corresponds to a certain direction and the width covers all 360 degrees. Each row corresponds to a different LiDAR beam. The values on the image reflect distance of points from the device. Using PGM representation of LiDAR inputs instead of BEV increased performance in \cite{sobh2018end}.

There is a variety of other LiDAR point cloud processing methods not yet explored in the end-to-end context \cite{lang2019pointpillars,yang2018pixor,zhou2018voxelnet,yan2018second,shi2019pointrcnn,wang2018sgpn}. More specialized reviews and comparisons of methods exist on this topic, in studies such as \cite{feng2019deep, zhanginstance}.

\subsection{High-definition maps}

The modular approach to autonomous driving relies heavily on accurate localization for route planning. High-definition (HD) maps are relatively costly to obtain, but for driving in dense urban environments they are imperative. In contrast, HD maps are not generally required for learning end-to-end policies -- just a camera input can be sufficient.

That said, HD maps can be incorporated into the end-to-end pipelines if needed. HD maps can provide an immense amount of information about the driving scene and make the task of self-driving a lot more manageable. Usage of HD maps in end-to-end driving displays a compromise between the two extremes of driving software stacks. Such approaches are sometimes referred to in the literature as \emph{mid-to-mid} \cite{bansal2018chauffeurnet}. 

\begin{figure}
    \centering
    \includegraphics[width=0.9\columnwidth]{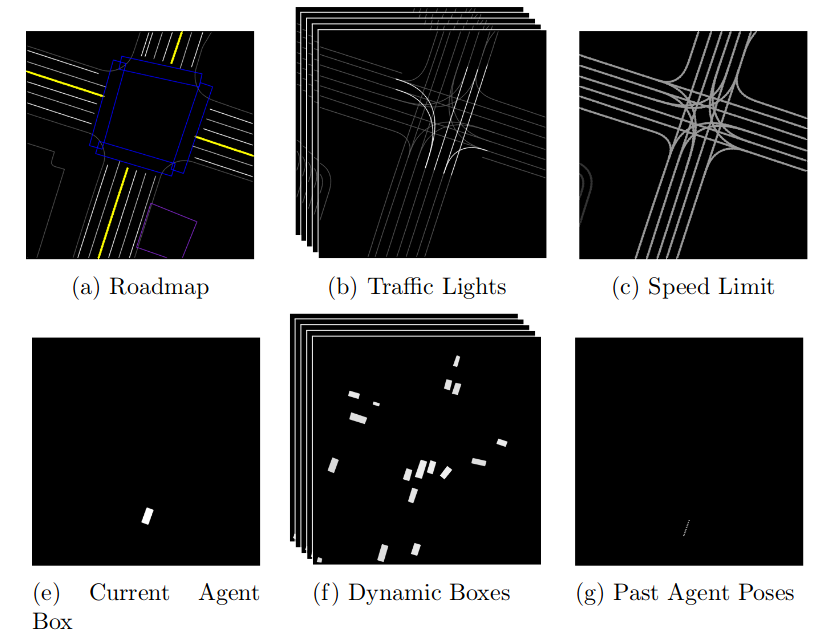}
    \caption{ChauffeurNet's \cite{bansal2018chauffeurnet} inputs are top-down HD maps containing both static information about the road and the locations of dynamic objects.}
    \label{fig:chauff_maps}
\end{figure}

In order to make use of HD maps in an end-to-end pipeline, it is common to put all the traffic information onto one or several top-down images in order to be processed by convolutional layers \cite{bansal2018chauffeurnet,zeng2019end,casas2018intentnet,chen2019learning}. Top-down HD maps may contain static information about roads, lanes, intersections, crossings, traffic signs, speed limits as well as dynamically changing information about traffic lights \cite{zeng2019end,casas2018intentnet,bansal2018chauffeurnet}. In addition to information about the road, ChauffeurNet \cite{bansal2018chauffeurnet} uses an existing perception module to detect and draw a map containing other agents in the driving scene (see Figure \ref{fig:chauff_maps}). 

Hecker \emph{et al.} \cite{hecker2019learning} state that while simpler approaches (e.g. camera only) allow to study relevant problems in self-driving, fully autonomous cars require the use of detailed maps. For this reason, they augment the Drive360 dataset from \cite{hecker2018end} with 15 measures (affordances) extracted from detailed maps (see Table \ref{tab:hecker_list}.)

\begin{table}[h]
\renewcommand{\arraystretch}{1.1}
\centering
\caption{List of measures (affordances) extracted from HERE Technologies detailed map and used as inputs in \cite{hecker2019learning}.}
\label{tab:hecker_list}
\begin{scriptsize}
    \begin{tabular}{|l|c|}
    \hline
        \textbf{Name and description} & \textbf{Range of values}\\
        \hline
        Road-distance to intersection & [0 m, 250 m]\\
        Road-distance to traffic light & [0 m, 250 m]\\
        Road-distance to pedestrian crossing & [0 m, 250 m]\\
        Road-distance to yield sign & [0 m, 250 m] \\
        \hline
        Legal speed limit & [0 km/h, 120 km/h] \\
        Average driving speed based on road geometry & [0 km/h, $\infty$ km/h] \\
        Curvature (inverse of radius) & [0 m$^{-1}$, $\infty$ m$^{-1}$]\\
        \hline
        Turn number: which way to turn in next intersection & [0,$\infty$]\\
        \hline
        Relative heading of the road after intersection & [-180$\deg$, 180 $\deg$]\\
        Relative heading of all other roads & [-180$\deg$, 180 $\deg$]\\
        \hline
        Relative headings of map-matched GPS coordinate  & [-180$\deg$, 180 $\deg$] x 5\\
        in \{1,5,10,20,50\} meters & \\
        \hline

    \end{tabular}
    \end{scriptsize}

\end{table}

\subsection{Multi-modal fusion}
In multi-modal approaches the information from multiple input sources must be combined \cite{feng2019deep}. The multiple modalities might be joined at different stages of computation:
\begin{itemize}
    \item Early fusion: multiple sources are combined before feeding them into the learnable end-to-end system. For example, one can concatenate RGB and depth images channel-wise. Pre-processing of inputs might be necessary before fusing (e.g. converting to the same reference frame, rescaling to match dimensions, etc.).
    \item Middle fusion: modalities are combined after some feature extraction is done on some or all of them. The feature extraction itself is most often part of the end-to-end model and is learnable. Further computations are performed on the joined (e.g. concatenated) features to reach the final output.
    \item Late fusion: outputs are calculated on each input modality separately. The separate outputs are then combined in some way. A well-known late-fusion approach is ensembling, for example using \emph{Kalman filters} \cite{houtekamer1998data} or \emph{mixture of experts} \cite{jacobs1991adaptive}.
\end{itemize}

In the works covered in this review we mainly encountered early and middle fusion approaches. Early fusion is often computationally the most efficient \cite{xiao2019multimodal}. The most common fusion technique is \emph{concatenation} that simply stacks inputs. Concatenating RGB color and depth channels proved to be the best-performing solution for merging RGB and depth inputs \cite{xiao2019multimodal}. Zhou \emph{et al.} \cite{zhou2019does} early-fused various semantic maps (segmentation, depth and others) with RGB images. In the case of concatenating LiDAR and visual inputs both early \cite{zeng2019end} and middle fusion \cite{casas2018intentnet,chen2018lidar,sobh2018end} have been successfully applied. Vehicle state measurements such as speed and acceleration are usually middle-fused with visual inputs \cite{codevilla2018end,liang2018cirl} (concatenation in both cases). Hecker \emph{et al.} \cite{hecker2018end,hecker2019learning} middle-fused visual temporal features obtained with long short-term memory (LSTM) \cite{hochreiter1997long} modules from camera feeds by concatenating them with features extracted from maps or from GPS coordinates. As an alternative fusion method, \emph{element-wise multiplication} was used by Kim \emph{et al.} \cite{kim2019grounding} to middle-fuse textual commands with visual information.

Late fusion has proved more efficient than early fusion in entering desired navigational command (go left, go right, go straight, continue) \cite{sauer2018conditional,codevilla2018end,xiao2019multimodal}. Specifically, navigational command is used to \emph{switch between output branches}. Alternatively, Chowdhuri \emph{et al.} \cite{chowdhuri2019multinet} solved a similar problem (switching behavior mode) with middle-fusion.

\subsection{Multiple timesteps}

Certain physical characteristics of the driving scene like speed and acceleration of self and other objects are not directly observable from a single camera image. It can therefore be beneficial to consider multiple past inputs via:
\begin{itemize}
    \item \textbf{CNN+RNN.} Most commonly CNN-based image processing layers are followed by a recurrent neural network (RNN, most often LSTM \cite{hochreiter1997long}). The RNN receives the sequence of extracted image features and produces final outputs \cite{xu2017end,chen2018lidar,bansal2018chauffeurnet}. Multiple RNNs can be used in parallel for performing different subtasks \cite{bansal2018chauffeurnet}. Also multiple sources of information (e.g. LiDAR and camera \cite{chen2018lidar}) can be processed by CNNs, concatenated and fed to the RNN together. Alternatively, spatio-temporal extraction can be done on each source separately, with the final output calculated based on concatenated RNN outputs \cite{hecker2018end,hecker2019learning}. Recurrent modules can also be stacked on top of each other -- Kim \emph{et al.} \cite{kim2019grounding} uses a CNN image encoder and a LSTM-based textual encoder followed by an LSTM control module.
    \item \textbf{Fixed window CNN.} Alternatively, a fixed number of previous inputs can be fed into the spatial feature extractor CNN module. The resulting feature maps can act as inputs for the succeeding task-specific layers \cite{sauer2018conditional}. For LiDAR, Zeng \emph{et al.} \cite{zeng2019end} stack the ten most recent LiDAR 3D occupancy grids along height axis and use 2D convolutions.    
\end{itemize} 

Driving is a dynamic task where temporal context matters. Using past information might help a very good driving model to remember the presence of objects that are momentarily obscured or to consider other drivers' behavior and driving styles.


\section{Output modalities}
\label{sec:outputs}
\subsection{Steering and speed}
The majority of end-to-end models yield as output the steering angle and speed (or acceleration and brake commands) for the next timestep \cite{pomerleau1989alvinn,hecker2018end,kim2017interpretable,codevilla2018end,liang2018cirl,sobh2018end}. Usually, this is treated as a regression problem, but by binning steering angles one can transform it into classification task \cite{xu2017end}. Steering wheel angle can be recorded directly from the car and is an easily obtained label for IL approaches. However, the function between steering wheel angle and the resulting turning radius depends on the car's geometry, making this measure specific to the car type used for recording. In contrast, predicting the inverse of the car's turning radius \cite{bojarski2016end} is independent of the car model’s geometry. Conveniently, the inverse turning radius does not go to infinity on a straight road. Notice that to achieve the desired speed and steering angle outputted by the network, additional PID controllers are needed to convert them to acceleration/brake and steering torque.

To enforce smoother driving, Hawke \emph{et al.} \cite{hawke2019urban} proposed to output not only the speed and the steering angle, but also the car's acceleration and angular acceleration of steering. Beyond smoothness, the authors also report better driving performance compared to just predicting values. Temporal consistency of output commands is also enforced in Hecker \emph{et al.} \cite{hecker2019learning} (see Section \ref{sec:safety}).

Many authors have recently optimized speed and steering commands using mean absolute error (MAE) instead of mean squared error (MSE) \cite{hecker2019learning,codevilla2019exploring,xiao2019multimodal,zhou2019does,sauer2018conditional,bansal2018chauffeurnet}. MAE has been shown to correlate better with actual driving performance \cite{codevilla2018offline}.

\subsection{Waypoints}
A higher-level output modality is predicting future waypoints or desired trajectories. Such approaches have been investigated, for instance, in \cite{bansal2018chauffeurnet,chowdhuri2019multinet,zeng2019end,caltagirone2017lidar,li2018oil}. The network's output waypoints can be transformed into low-level steering and acceleration commands by another trainable network model, as in \cite{li2018oil}, or via a controller module, for instance as in \cite{bansal2018chauffeurnet, chen2019learning}. Different control algorithms are available for such controller modules, a popular one being PID. Such controller modules can reliably generate the low-level steering and acceleration/braking commands to reach the desired points. For smoother driving, one can fit a curve to the noisy predicted waypoints and use this curve as the desired trajectory \cite{chen2019learning}.

In contrast to predicting momentary steering and acceleration commands, by outputting a sequence of waypoints or a trajectory, the model is forced to plan ahead. A further advantage of using waypoints as output is that they are independent of car geometry. Furthermore, waypoint-based trajectories are easier to interpret and analyze than low-level network outputs such as momentary steering commands.

\subsection{Cost maps}

\begin{figure}
    \centering
    \includegraphics[width=\columnwidth]{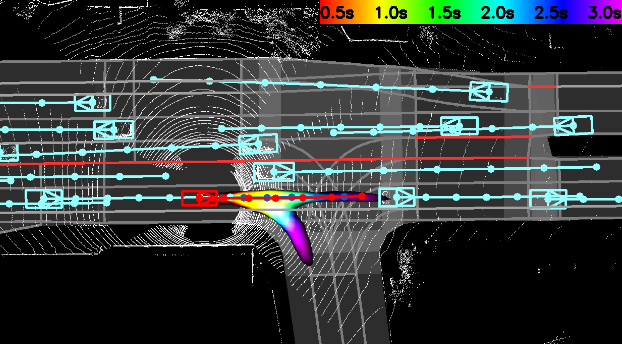}
    \caption{Cost Volume across Time produced by Zeng \emph{et al.} \cite{zeng2019end}. The planned trajectory is shown as a red line, ground-truth (human) trajectory as a blue line. The lowest cost regions for a number of future timesteps are overlaid using different colors (each color represents the low-cost region for a separate timestep, indicated by the legend). Detection of other objects in the driving scene and corresponding motion prediction results are in cyan. Figure adapted from \cite{zeng2019end}}.
    \label{fig:zeng_cost}
\end{figure}

In many cases a variety of paths are equally valid and safe. It has been proposed to output cost maps, which contain information about where it is safe to drive \cite{drews2017aggressive, zeng2019end}. The cost maps are then used to pick a good trajectory and model predictive control (MPC) \cite{FALCONE_TCST_2007} computes the necessary low-level commands. \cite{zeng2019end} produced top-down view 2D cost maps for a number of future time steps based on LiDAR and HD maps. Human expert trajectories were used for training the cost map prediction by minimizing human trajectory cost and maximizing random trajectory cost. Based on this cost volume, potential trajectories were evaluated (see Figure \ref{fig:zeng_cost}). Visualizing the cost maps allows humans to better understand the machine’s decisions and reasoning.

\subsection{Direct perception and affordances}

\emph{Direct perception} \cite{chen2015deepdriving} approaches aim to fall between modular pipelines and end-to-end driving and combine the benefits of both approaches. Instead of parsing all the objects in the driving scene and performing robust localization (as modular approach), the system focuses on a small set of crucial indicators, called \emph{affordances} \cite{chen2015deepdriving,sauer2018conditional,al2017deep,seff2016learning}. For example, the car’s position with respect to the lane, position relative to the edges of the road and the distances to surrounding cars can be predicted from inputs \cite{chen2015deepdriving}. The affordance values can be fed to a planning algorithm to generate low-level commands for safe and smooth driving. 

\begin{figure}[h]
    \centering
    \includegraphics[width=0.8\columnwidth]{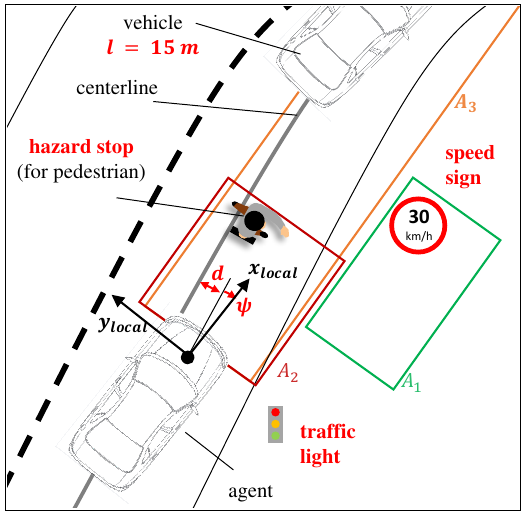}\\
    \setlength{\tabcolsep}{3pt}

    \begin{scriptsize}
    \begin{tabular}{clcl}
    \hline
    Conditional & Affordances & Acronym  & Range of values\\
    \hline
    No & Hazard stop &	-	   & $\in \{True, False\}$ \\ 
    & Red Traffic Light              &  -       & $\in \{True, False\}$      \\ 
    & Speed Sign {[}km/h{]}          & 	-	   & $\in \{None, 30, 60, 90\}$ \\
    \hline
    No & Distance to vehicle {[}m{]}    & 	$\ell$	   & $\in [0, 50]$ 	\\ 
    \hline
    Yes & Relative angle {[}rad{]}       &  $\psi$	   & $\in [-\pi, \pi]$\\ 
    & Distance to centerline {[}m{]} &	$d$	   & $\in [-2, 2]$              \\
    \hline
    \end{tabular}
    \end{scriptsize}

    \caption{Affordances used by Sauer \emph{et al.} \cite{sauer2018conditional}. \textbf{Top:} Illustration of the affordances (red) and observation areas $A_1$ to $A_3$ used by the model. Within $A_1$ traffic lights and speed signs are detected. If there is an obstacle in $A_2$ hazard stop label is set to True and the agent is expected to stop. Distance to vehicle ahead is measured in $A_3$. \textbf{Bottom:} List of the affordances. They can be discrete or continuous and either conditional (dependent on directional  input) or not. }
    \label{fig:sauer_aff}
\end{figure}

\emph{Direct perception via affordances} was combined with \emph{conditional driving} in \cite{sauer2018conditional} by providing high-level navigational commands. Furthermore, the predicted affordances in \cite{sauer2018conditional} include speed signs, traffic lights and unexpected traffic scene agents (see Figure \ref{fig:sauer_aff}).
 
\subsection{Multitask learning}

In multitask learning \cite{vapnik2009new}, multiple outputs are predicted and optimized simultaneously. For example, in addition to planning the trajectory of self, driving models can be asked to also detect and predict the motion of other objects on the scene \cite{bansal2018chauffeurnet,zeng2019end}. These additional outputs are computed by a separate branch based on an intermediate layer of the network. While these tasks are only indirectly relevant for the driving task (the main task), they provide a rich source of additional information that conditions the internal representations of the model in the shared layers. Results show that simultaneously optimizing such side-tasks is beneficial and results in a more robust model \cite{bansal2018chauffeurnet,zeng2019end,xu2017end}. 

As an additional benefit, the additional outputs can potentially help comprehend the driving decisions and failures of the end-to-end model (more details in Section \ref{sec:interpretability}).

\section{Evaluation}
\label{sec:evaluation}

It is not always cost-effective to test each model thoroughly in real life. Deploying each incremental improvement to measure its effect is costly. Deploying experimental approaches in real traffic is outright dangerous. In the case of a modular approach, it is common to evaluate each module independently against benchmark datasets such as KITTI \cite{geiger2013vision} for object detection. Nevertheless, performance in sub-tasks does not directly relate to actual driving performance and it is not applicable to end-to-end models as there are no intermediate outputs to evaluate. Hence, a different set of metrics has been developed for end-to-end approaches.

\begin{figure}
    \centering
    \includegraphics[width=\columnwidth]{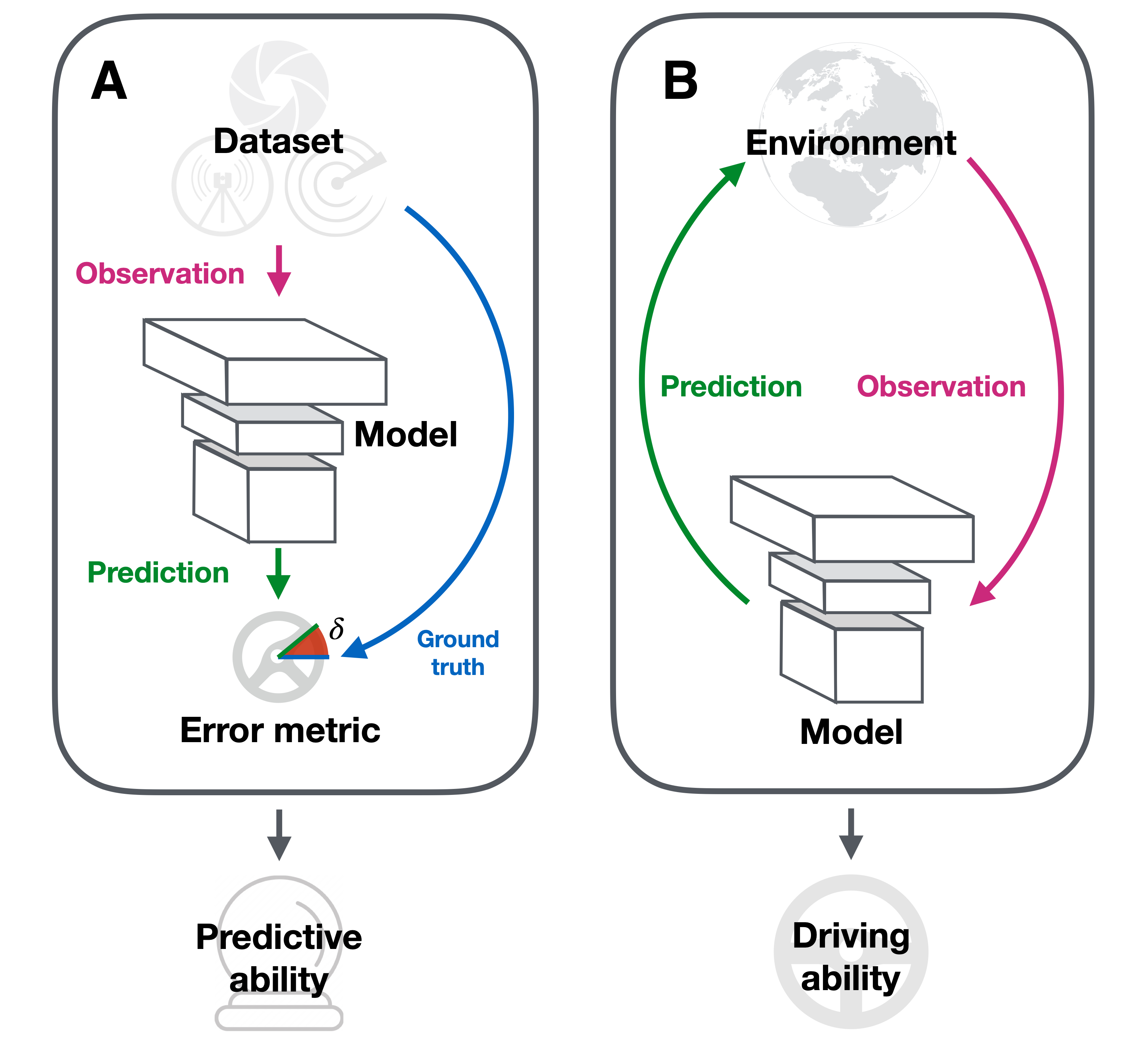}
    \caption{Open and closed-loop evaluation. \textbf{A:} Open-loop evaluation uses a part of the original dataset to evaluate the model. Similarity between predicted and ground truth values is measured. No actual driving is done. \textbf{B:} Closed-loop evaluation deploys the model in an environment. The model's predictions are used as driving actions. The resulting driving behavior is observed and quantified.}
    \label{fig:open_closed}
\end{figure}

The easiest way to test imitation learning models is \emph{open-loop evaluation} (Figure \ref{fig:open_closed} left). In such evaluation, the decisions of the autonomous-driving model are compared with the recorded decisions of a human driver. Typically, a dataset is split into training and testing data. The trained model is evaluated on the test set by some performance metric. The most common metrics are mean absolute error and mean squared error of network outputs (e.g. steering angle, speed). A more thorough list of open-loop metrics is given in Table \ref{tab:offline}. Note that there might be multiple equally correct ways to behave in each situation and similarity with just one expert's style might not be a fair measure of driving ability. Particularly in the case of RL agents, the models may come up with safe driving policies that are not human-like according to mean errors. Hence, open-loop evaluation is restricted to IL and is not used for models trained with RL. Despite its limitations, some end-to-end models are only open-loop evaluated \cite{hecker2018end,zeng2019end,amini2019variational,hecker2019learning,pan2017virtual,xu2017end,kim2019grounding}.

\begin{table}[]
\centering
\caption{Open-loop evaluation metrics used in \cite{codevilla2018offline}. $a$ is the continuous ground-truth action, $\hat{a}$ is the predicted action, $V$ is a set of samples used in validation, $\delta$ is the Kronecker delta function, $\theta$ is the Heaviside step function, $Q$ is a quantization function. }
\renewcommand{\arraystretch}{1.1}
\setlength{\tabcolsep}{3pt}
\begin{scriptsize}
\begin{tabular}{p{30mm}cl}
  Metric name                               & Parameters & Metric definition\\
  \hline
  Squared error                             & --          & $\frac{1}{|V|}\sum\limits_{i \in V} \| a_i - \hat{a}_i \|^2$ \\
  Absolute error                            & --          & $\frac{1}{|V|}\sum\limits_{i \in V} \| a_i - \hat{a}_i \|_1$ \\
  Speed-weighted absolute error             & --          & $\frac{1}{|V|}\sum\limits_{i \in V} \| a_i - \hat{a}_i \|_1 v_i$ \\
  Cumulative speed-weighted absolute error & $T$        & $\frac{1}{|V|}\sum\limits_{i \in V} \| \sum\limits_{t=0}^T (a_{i+t} - \hat{a}_{i+t}) v_{i+t} \|_1$ \\
  Quantized classification error            & $\sigma$   & $\frac{1}{|V|}\sum\limits_{i \in V} \left(1-\delta \left(Q(a_i, \sigma), Q(\hat{a}_i, \sigma) \right)\right)$ \\
  Thresholded relative error                & $\alpha$   & $\frac{1}{|V|}\sum\limits_{i \in V} \theta\left(\|\hat{a}_i - a_i\| - \alpha \|a_i \| \right)$ \\
  \hline
\end{tabular}
\end{scriptsize}
\label{tab:offline}
\end{table}
In contrast, \emph{closed-loop evaluation} directly evaluates the performance of a driving model in a realistic (real or simulated) driving scenario by giving the model control over the car (Figure \ref{fig:open_closed} right). Unlike in the real world, in simulation closed-loop evaluation is easy and not costly to perform. Furthermore two benchmarks exist for the CARLA simulator, allowing a fair comparison of models \cite{dosovitskiy2017carla,codevilla2019exploring}. Despite the inherent danger, real-life closed-loop testing has also been reported for end-to-end models \cite{bewley2019learning,hawke2019urban,kendall2019learning,bansal2018chauffeurnet,bojarski2016end,codevilla2018end}. The closed-loop metrics used in literature include:
\begin{itemize}
    \item percentage of successful trials, \cite{dosovitskiy2017carla, codevilla2019exploring,hawke2019urban,bansal2018chauffeurnet,liang2018cirl,muller2018driving,sauer2018conditional,xiao2019multimodal,zhou2019does}
    \item number of infractions (collisions, missed turns, going off road, etc), \cite{muller2018driving,sobh2018end,codevilla2018end},
    \item average distance between infractions  \cite{dosovitskiy2017carla,bewley2019learning,codevilla2019exploring,kendall2019learning} or disengagements \cite{dmv2019reports} ,
    \item time spent on lane markings or off-road, \cite{sobh2018end}
    \item percentage of autonomy (i.e. percentage of time car is controlled by the model, not safety driver), \cite{bojarski2016end},
    \item fraction of distance travelled towards the goal \cite{codevilla2018offline}.
\end{itemize} 
Clearly, all these metrics directly measure the model's ability to drive on its own, unlike open-loop metrics.

Indeed, good open-loop performance does not necessarily lead to good driving ability in closed-loop settings. Codevilla \emph{et al.} \cite{codevilla2018offline} performed extensive experimentation (using 45 different models) to measure correlations between different open-loop and closed-loop metrics. The used open-loop metrics are listed in Figure \ref{tab:offline}, the closed-loop metrics were 1) trial success rate, 2) fraction of distance traveled towards goal, and 3) average distance between infractions. The results showed that even the best offline metrics only loosely predict closed-loop performance. Mean squared error correlates with closed-loop success rate only weakly (correlation coefficient $r=0.39$), so mean absolute error, quantized classification error or thresholded relative error should be used instead ($r>0.6$ for all three).
Beyond these three suggested open-loop measures, \emph{balanced-MAE} was recently reported to correlate better with closed-loop performance than simple MAE \cite{bewley2019learning}. Balanced-MAE is computed by averaging the mean values of unequal-length bins (according to steering angle). Because most data lies in the region around steering angle 0, equally weighting the bins grows the importance of rarely occurring higher steering angles.

While open-loop metrics are not sufficient to measure driving ability, when experimenting with various models, hyper-parameter tuning or when deciding which loss function to use, it is useful to know that some open-loop metrics correlate better with eventual driving ability than others. For example, Bewley \emph{et al.} \cite{bewley2019learning} used balanced-MAE to select the best models to test in closed-loop.

Beyond just measuring ability to drive without infractions, Hecker \emph{et al.} \cite{hecker2019learning} proposed to measure the human-likeness of behavior using generative adversarial networks. More human-like driving is argued to be more comfortable and safer. For measuring comfort, one could also evaluate models according to longitudinal and lateral jerk that are major causes of discomfort (as done in the Supplementary Information of \cite{sauer2018conditional}).

Undoubtedly, the most relevant measure of quality of self-driving models is the ability to drive without accidents in real traffic. For comparing the safety of autonomous driving solutions, the State of California requires manufacturers to report each traffic collision involving an autonomous vehicle. These reports are publicly available\footnote{\url{https://www.dmv.ca.gov/portal/dmv/detail/vr/autonomous/autonomousveh_ol316}}. Furthermore, manufacturers testing autonomous vehicles on public roads also submit an annual report summarizing the disengagements (interventions of safety driver) during testing\footnote{\url{https://www.dmv.ca.gov/portal/dmv/detail/vr/autonomous/testing}}. These reports allow to compare the performance of different self-driving technologies by miles-per-disengagement and miles-per-accident metrics, i.e. the real-life performance. The shortcomings of these measures are examined in Discussion (Section \ref{sec:discussion}).

\section{Interpretability}
\label{sec:interpretability}

In case of failures, it is crucial to understand why the model drives the way it does, so that similar failures could be avoided in the future. While neural networks perform highly complex hierarchical computations, certain methods allow us to investigate what is happening inside the models.

\subsection{Visual saliency}

A whole body of research exists on how to interpret the computations that convolutional neural networks perform on visual inputs \cite{zeiler2014visualizing,simonyan2013deep,bach2015pixel,dabkowski2017real}. \emph{Sensitivity analysis} aims to determine the parts of an input that a model is most sensitive to. The most common approach involves computing the gradients with respect to the input and using the magnitude as the measure of sensitivity. While popular in other domains and not restrictively expensive computationally, there is no notable application of gradient-based approach in end-to-end driving.

\begin{figure}
    \centering
    \includegraphics[width=\columnwidth]{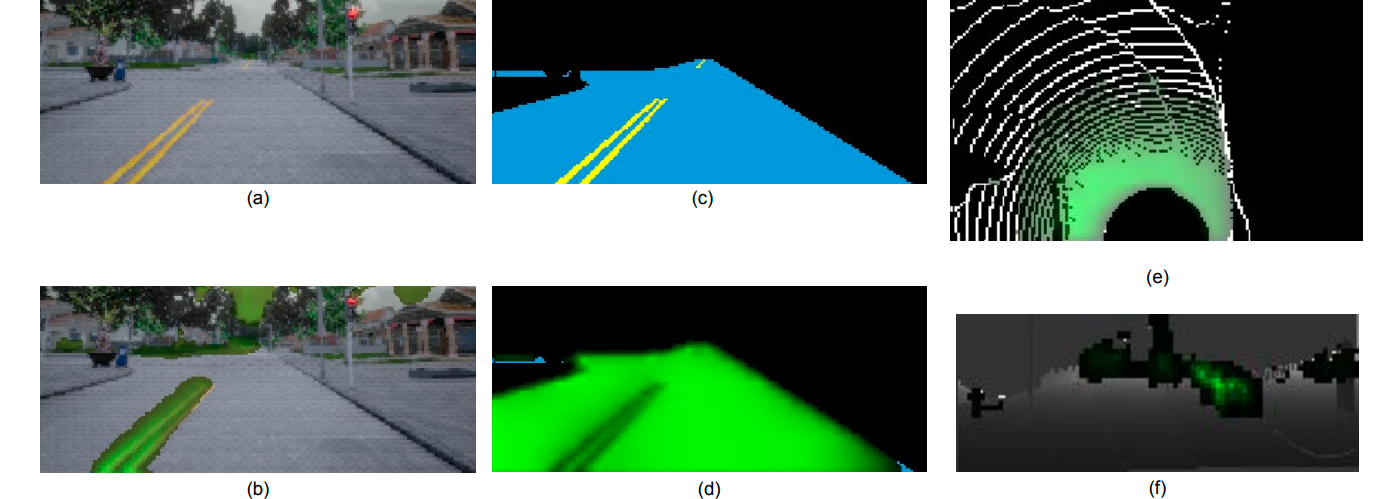}
    \caption{Saliency for different input types as illustrated in \cite{sobh2018end}. Salient parts of inputs are colored green. (a) Original RGB camera image; (b) Attention map overlaid on the Original RGB image;
(c) Ground truth semantic segmentation; (d) Attention map overlaid on the semantically segmented
image; (e) Attention map overlaid on LiDAR bird view image; (f) Attention map overlaid on
LiDAR data with PGM processing }
    \label{fig:salient}
\end{figure}

The VisualBackProp method \cite{bojarski2018visualbackprop} is a computationally efficient heuristic to determine which input pixels influence the car’s driving decision the most \cite{bojarski2017explaining}. Shifting the salient regions of images influences steering prediction linearly and almost as much as shifting the whole image, confirming they are relevant. VisualBackProp can also be applied to other types of inputs, such as segmented images and 2D projections of LiDAR point clouds \cite{sobh2018end} (see Figure \ref{fig:salient}). 

\begin{figure}
    \centering
    \includegraphics[width=\columnwidth]{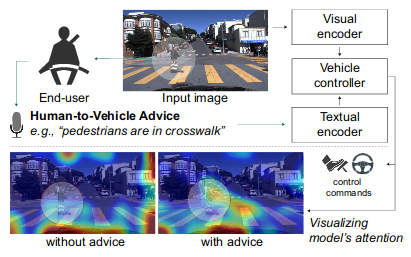}
    \caption{Model in \cite{kim2019grounding} can take in consideration textual commands from the user. These commands influence which regions of the inputs the model pays more attention to.}
    \label{fig:attention}
\end{figure}

While gradient-based sensitivity analysis and VisualBackProp highlight important image regions for an already trained model, \emph{visual attention} is a built-in mechanism present already when learning. The model predicts a spatial mask of weights, corresponding to ``where to attend'', which is then used to scale inputs. As feature maps preserve spatiality, visual attention can be applied on the feature maps from the convolutional layers \cite{kim2017interpretable,kim2019grounding}. Where to attend in the next timestep (the \emph{attention mask}), is predicted as additional output in the current step and can be made to depend on additional sources of information (e.g. textual commands \cite{kim2019grounding}). Masks can be upsampled and displayed on the original image (see Figure \ref{fig:attention}).

\subsection{Intermediate representations}

Interpretability is a definite advantage of the engineered intermediate representations used in modular pipelines. 
End-to-end models can also benefit from semantic representations as inputs to the model. In addition to improving model generalization \cite{zhou2019does} they allow to separately investigate the errors made in predicting the semantic images (segmentation, depth image etc.) and the errors made by the driving model built on them.

\emph{Direct perception} approaches \cite{sauer2018conditional,chen2015deepdriving,al2017deep} predict human-understandable affordances that can then be used by hard-coded controllers to make detailed driving decisions. Similarly, waypoints are quite intuitively understandable. Outputting waypoints or affordances instead of the low-level driving commands adds interpretability and helps directly pinpoint the reason (e.g. error in a certain affordance) that lead to a questionable driving decision. Similarly, failures of driving scene understanding are clearly visible when visualizing cost maps that are the outputs of the learnable part of the pipeline in \cite{zeng2019end} and \cite{drews2017aggressive}. The cost maps can be transformed into actions via path planning and control methods.

\subsection{Auxilliary outputs}
While one can visualize the sequence of commands or predicted routes, how the system reached these outputs is often not clear. Designing the model to simultaneously predict additional outputs (i.e. \emph{auxilliary outputs}) and can help comprehend the driving decisions, a benefit beyond the main goal of helping the model to learn more efficient representations 

The main and side-tasks (auxilliary tasks) rely on the same intermediate representations within the network. The tasks can be optimised simultaneously, using multiple loss functions. For example, based on the same extracted visual features that are fed to the decision-making branch (main task), one can also predict ego-speed, drivable area on the scene, and positions and speeds of other objects \cite{bansal2018chauffeurnet,zeng2019end,codevilla2019exploring}. Such auxilliary tasks can help the model to learn better representations via additional learning signals and via constraining the learned representations \cite{bansal2018chauffeurnet,codevilla2019exploring,zeng2019end}, but can also help understand the mistakes a model makes. A failure in an auxilliary task (e.g. object detection) might suggest that necessary information was not present already in the intermediate representations (layers) that it shared with the main task. Hence, also the main task did not have access to this information and might have failed for the same reason.  

\section{Safety and comfort}
     
\label{sec:safety}
Safety and comfort of passengers are prerequisites of commercial applications of self-driving cars. However, millions of hours of driving are needed to prove a reasonable-sized fleet of cars causes accidents rarely enough \cite{kalra2016driving,norden2019efficient}. Also, the diversity of test driving data is almost never sufficiently high to ensure the safety of driving models in all conditions \cite{pei2017deepxplore}. For example, NVIDIA DAVE-2 \cite{bojarski2016end} can be made to fail by just changing the light conditions \cite{pei2017deepxplore}. Similarly, many more recent and more advanced models fail to generalize to combinations of new location and new weather conditions, especially in dense traffic, even in simulation \cite{codevilla2019exploring}. 

It is known that neural networks can be fooled with carefully designed inputs, especially if one has access to the model parameters (white box attacks) \cite{szegedy2013intriguing,goodfellow2014explaining}. For example, it has been shown that perception models can be fooled by placing stickers on traffic signs \cite{eykholt2018robust}. Also, driving assistance systems can be fooled by putting stickers or projecting images on the road \cite{nassi2020phantom}. There is no evident reason why end-to-end driving models should by default be robust to such attacks, hence before the technology can be deployed specific measures need to be taken to avoid this vulnerability.

The real world is expected to be even more diverse and challenging for generalization than simulation. Testing enough to get sufficient statistical power is hard and expensive to achieve, and also puts other traffic participants in danger \cite{norden2019efficient}. Hence simulations are still seen as the main safety-testing environment \cite{norden2019efficient, o2018scalable}. In simulation one can make the model play through many short (few to tens of seconds) periods of critical, dangerous situations \cite{norden2019efficient}. In real life each play-through of such situation would need the involvement of many people and cars and resetting the scene each time.

One can avoid the need to generalize to all cases if the model can reliably detect its inability to deal with the situation and pass the driving on to a person \cite{hecker2018failure} or to another algorithm. This means adding a safety module detecting situations where the self-driving model is likely to fail \cite{hecker2018failure}. Also Bayesian approaches can be used to estimate the uncertainty of model outputs, as proposed in multiple recent works \cite{lee2019early,filos2020can}.

Ensuring the comfort of passengers is a distinct problem from safety. Comfort includes motion comfort (sickness), apparent safety and level of controllability \cite{kuderer2015learning,hecker2019learning}. Motion comfort can be increased by reducing longitudinal and lateral jerk via an additional loss function enforcing temporal smoothness of steering and speed \cite{hecker2019learning}. Note that for passenger comfort speed and steering should depend on each other (as in \cite{attia2014combined,han2010bezier}), instead of being two independent outputs. While not mentioning comfort as a desired outcome, ChauffeurNet \cite{bansal2018chauffeurnet} included an additional loss function (\emph{geometry loss}) to enforce smooth trajectories.

Human-like driving increases safety and comfort as people both inside and outside the vehicle can better comprehend the model's driving behavior. An adversarial network attempting to discriminate between human and machine driving was added in \cite{hecker2019learning}. The adversarial loss increased the comfort as measured by the lateral and longitudinal jerk, and achieved improved accuracy over pure imitation learning.

\section{Discussion}
\label{sec:discussion}

In this section we attempt to summarize the best, most promising practices in end-to-end driving.

\subsection{Architectures}

There is a huge diversity of architectures that could be used in end-to-end driving. In here we try to narrow down the space of likely to be useful models by discussing the most promising choices for inputs, outputs, fusion techniques and auxilliary tasks.

Driving based on only camera inputs makes the eventual deployment of the technology affordable. Measurements such as current speed and acceleration are also easy to obtain. However, using LiDAR and HD maps puts end-to-end models in the same price range with modular approaches, making it not affordable for everyone. The initial cost of sensors, the cost of creating and maintaining HD maps and the cost of repairing and insuring a car with many sensors\cite{wired2020new}  prevent the wider adoption of LiDAR and map-based approaches. We believe that pursuing affordable self-driving based on the end-to-end approach is of more interest for car manufacturers. This is opposed to ride-hailing service providers and trucking companies, whose cost model can accommodate the increased price due to the self-driving technology. 

A 360-degree view around the vehicle is shown to be useful for more complicated driving maneuvers like lane changing or giving way on intersections. Conversely, stereo-vision for depth estimation is not commonly used, probably because it is useful only in the close proximity (10m) \cite{strat1992natural}.

Another important input modality is navigational instructions. Providing the route as a map image is the more flexible option, as it defines the intended route more precisely in a longer time-scale. In contrast, with categorical commands, the instruction "turn left" might be confusing or come too late if there are multiple roads to turn into on the left. An average human driver would also find it hard to navigate in a foreign city based on only the voice instructions of a navigation app. 
We hence conclude that while it is not clear how well the model is capable of extracting route information from the route planner screen image, this approach is more flexible and more promising in longer term.

Models usually have multiple inputs. With several input sources (e.g. multiple cameras, self-speed, navigation input), one needs to merge the information in some way. Early fusion is appealing as all pieces of information can be combined from early on. One could early-fuse maps, visual inputs and LiDAR data for example by concatenating them as different channels, but for this they must be mapped to the same frame of reference and be of equal spatial size. Early-fusion seems hard to apply for inserting speed and other non-spatial measurements into the model. Hence, middle-fusion remains the default strategy that can be applied to all inputs.

It is beneficial to endow a model with computer-vision capabilities. Zhou \emph{et al.} \cite{zhou2019does} propose training a separate set of networks to predict semantic segmentation, optical flow, depth and other human-understandable representations from the camera feed. These images can then be used as \emph{additional inputs} (e.g. early-fused by stacking as channels). 
Alternatively, the these semantic images can be used as targets for auxiliary tasks. In multi-task learning, the main branch of the model transforms images into driving outputs, while at certain layer(s) the model forks out to produce additional outputs such as segmentation maps and depth maps. 
In comparison to generating and using additional inputs, the approach of using auxiliary tasks has one major benefit - the specialized networks (and the branches) are not needed at execution time, making the eventual deployed system computationally faster. On the other hand, using those additional branches during evaluation allows to reason about the mistakes the network made - if the model did not detect an object it may be the reason why it did not avoid it.

The outputs of the model define the level of understanding the model is expected to achieve. When predicting instantaneous low-level commands, we are not explicitly forcing the model to plan a long-term trajectory. Ability to plan ahead might arise in the internal representations, but it is neither guaranteed nor easily measurable. When predicting a series of future desired locations (e.g. waypoints), the model is explicitly asked to plan ahead. Outputting waypoints also increases interpretability, as they can be easily visualized. Model outputs can be noisy, so waypoints should be further smoothed by fitting a curve to them and using this curve as the desired trajectory, as done in \cite{chen2019learning}. Speed can be deduced from distances between consecutive waypoints, but can also be an explicit additional output, as done in ChauffeurNet \cite{bansal2018chauffeurnet}. If deemed necessary, additional constraints can be added, such as forcing human-like trajectories (see Hecker \emph{et al.} \cite{hecker2019learning}). 

Outputting a series of cost-maps is an equally promising approach, allowing to plan motion many seconds ahead. The model can be made to estimate not only the instantaneous cost map, but also the cost maps for multiple future timesteps. A planner can then generate candidate paths and select a trajectory that minimizes the cost over multiple future time points.

\subsection{Learning}

Imitation learning is the dominant strategy for end-to-end autonomous driving. A list of existing datasets for IL has been provided in Appendix B of this survey. Despite multiple large datasets being available, it is common for authors to collect their own data. It is hard to disentangle if differences in results are due to the architecture or training data, even if benchmarks exist. Furthermore, network training and inference times are often not reported in the reviewed literature. This is unfortunate, as it would allow the readers to estimate the hardware requirements for training and to evaluate the models for real-time usage. Therefore, we encourage researchers to report this aspect in their future works


Online training \cite{chen2019learning} allows to avoid the distribution shift problem commonly associated with imitation learning. The case where the supervision signal can be queried in any state of the world and for any possible navigation command 
is particularly promising. Indeed, such an online-trained vision-only agent performed remarkably well in CARLA and NoCrash benchmark tasks \cite{chen2019learning}. Once trained, the vision-only model no longer needs the expert nor the detailed inputs. In the real world, however, creating the necessary expert policy (that can be queried in any state) is complicated, as there are only few companies reporting required level of performance from their modular driving stacks. Advances in sim2real may help such online-trained model to generalize from the simulation to the real world.

Recently, \cite{codevilla2019exploring} reported that using more training data from CARLA Town1 decreases generalization ability in Town2. This illustrates that more data without more diversity is not useful. The not-diverse datapoints contain the same information over and over again and lead to overfitting. As a potential remedy, one should weight the rare outputs and rare inputs (rare situations, locations, visual aspects, etc.) higher. The error the model makes on a datapoint might be a reasonable proxy for datapoint novelty. One could sample difficult data more frequently (as in prioritized experience replay used in RL \cite{schaul2015prioritized}) or weight difficult samples higher. This approach promises to boost learning in the long tails of both the input and output distributions.

Augmenting the data also improves the generalization ability. Common techniques such as blurring inputs, dropping out pixels, perturbing colors, adding noise or changing light conditions are known to work for standard supervised learning tasks, therefore can be expected to be beneficial also for imitation learning. Adding temporally correlated noise to the control commands during data collection is a common method for diversifying data that improves the generalization performance of imitation learning.

\subsection{Evaluation}
Off-policy imitation learning builds models via maximizing an open-loop performance, while actually the model is deployed in closed-loop settings. There are open-loop metrics that correlate better with closed-loop performance and therefore should be preferred during training. For example MAE is shown to be advantageous to MSE in that sense \cite{codevilla2018offline}. Furthermore, Bewley \emph{et al.} \cite{bewley2019learning} reported Balanced-MAE correlating even stronger with driving ability, which suggests training set balancing being also important for closed-loop performance.

One of the goals of evaluation is to give a fair estimation of the generalization ability of the model, i.e. the expected performance in new situations. 
Since the CARLA benchmark was released and adopted, multiple works have shown in simulation that model performance does not degrade drastically in a new city and unseen weather conditions \cite{codevilla2019exploring,chen2019learning}. However, the generalization ability drops sharply when increasing traffic density. 

While for CARLA-based models two benchmarks exist, models trained and tested in other simulators or on real world data have no clear comparison baselines. So an author can show high performance by choosing to test in simple settings. Therefore, readers should always pay close attention to the testing conditions of models.

The problem of comparing performance metrics obtained in different difficulty levels also applies for the safety measures collected by the State of California\footnote{\url{https://www.dmv.ca.gov/portal/dmv/detail/vr/autonomous/testing}} - miles per disengagement or miles per accident do not reveal where and in which conditions these miles were driven. Furthermore, it is not commonly agreed what constitutes a disengagement and different companies might apply different thresholds. There is a need for universal real-life evaluation procedure that would make a wide variety of different approaches comparable.

For allowing to compare different end-to-end models, future models should perform closed-loop evaluation: 
\begin{itemize}
    \item In diverse locations. Locations not used during training, if possible.
    \item In diverse weather and light conditions. Conditions not used during training, if possible. 
    \item In diverse traffic situations, with low, regular or dense traffic.
\end{itemize}
If training the model in CARLA simulator, one should report the performance in CARLA and NoCrash benchmarks. 

\subsection{Candidate architecture}
\begin{figure}[t]
    \centering
    \includegraphics[width=\columnwidth]{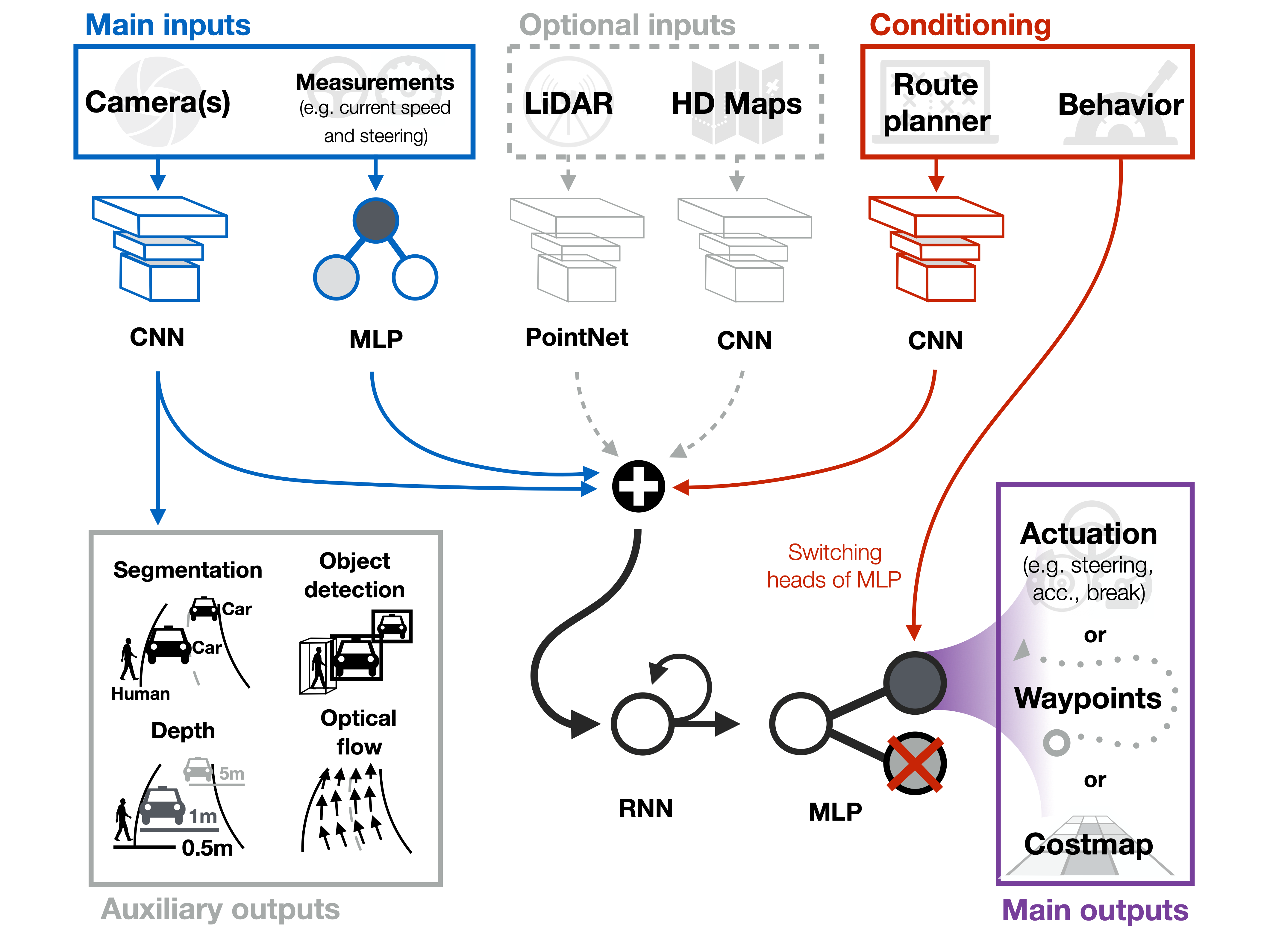}
    \caption{Candidate architecture summarizing the reviewed papers. \emph{Top:} the network receives input from multiple sensors. The inputs that are costly to use in the real world are marked optional. Navigational conditioning can be achieved via behavior command or route planner screen image. Behavior represents a categorical command, e.g. "turn left", "change lane right" or "drive carefully". Route planner screen image fulfills a similar role, but contains also extra information about the surrounding road network and buildings. \emph{Center:} the inputs are processed with convolutional or fully-connected layers and fused via concatenation (+). A RNN module extracts temporal information from a sequence of these fused representations and a fully-connected module calculates the final driving outputs.  \emph{Bottom right:} the final outputs are usually either actuation values, waypoints or cost maps. In the case of conditioning with behavioral command, the network has multiple sets of output nodes, one set per one categorical command. Which set of output nodes is used, is determined by switching according to the command. \emph{Bottom left:} jointly optimizing the main task and the auxilliary tasks shapes the internal representation of the camera CNN to include more semantic information. Additional auxilliary tasks can be added, for example predicting speed based on visual inputs or 3D object detection based on LiDAR.
    }
    \label{fig:architecture}
\end{figure}

Based on the most promising approaches in the end-to-end models reviewed in this survey article, we propose a candidate architecture to visualize what a modern end-to-end model looks like. This architecture has not been implemented and should be taken as an illustration. Such illustration can help readers to more intuitively grasp the structure of end-to-end models. The candidate architecture is given in Figure \ref{fig:architecture}.

{}

\subsection{Future challenges}

To drive safely in the real world, an autonomous driving model needs to handle smoothly the long tail of rare traffic situations. Commonly, the collected datasets contain large amounts of repeated traffic situations and only few of those rare events. Naive training on the whole dataset results in weak performance on the rare situations, because the model optimizes for the average performance to which the frequent situations contribute more. To put more emphasis on atypical situations during optimization, it has been proposed to balance the dataset according to the output distribution - i.e. making the rarely occurring driving commands more influential. However, this solves only part of the problem, because situations with atypical inputs, but common outputs, also exist - e.g. driving straight with a steady speed in rare weather conditions. Furthermore, it is in fact the joint distribution of inputs and outputs that defines the rarity of a data point. It is therefore crucial to develop data sampling techniques aiding the model to learn to deal with rare data points in a more general fashion.

At the same time the ability to deal with unusual situations is very hard to reliably test in the real world. The huge number of possible rare events means that billions of miles need to be driven before one can statistically claim an autonomous driving model is safe \cite{norden2019efficient}. Such \emph{black-box testing}, i.e. testing the entire driving pipeline together, is the only evaluation option for end-to-end models, which cannot be divided into separately verifiable sub-modules. Evaluation by deploying the entire model is also unavoidable for advanced modular approaches, where the interconnections between modules are complex and small errors may be amplified and cause unexpected behavior. Hence, the problem of black-box testing is universally relevant for all autonomous driving models and has been discussed intensively over the years \cite{kalra2016driving,koopman2016challenges, tian2018deeptest, o2018scalable, norden2019efficient}. There is a need for testing methods that would have safety guarantees for the real world.

Testing in simulation can be seen as one solution to the high costs of real-world testing. However the ability to pass simulated scenarios does not directly translate into safe driving in the real world. Further advances in making simulations more realistic or the development of better domain adaptation techniques are needed. One particularly challenging area is the modeling of the behavior of other traffic participants, needed for life-like simulations.

\appendices

\begin{landscape}
\section{Recent contributions in end-to-end driving}
\label{sec:contributions}

\setcounter{footnote}{0}
\begin{table}[h]
\begin{tiny}
\setlength{\tabcolsep}{2pt}
\caption{Recent contributions in end-to-end driving. The included articles are mainly selected to either test the model in real life or perform closed-loop testing in simulation. Abbreviations introduced in this table: cross-entropy (CE), Gaussian mixture model (GMM).}

\label{tab:articles}
\centering

\begin{tabular}{|c|c|c|c|c|c|c|}
\hline
ARTICLE & DATA TYPE & INPUTS & OUTPUTS & LOSSES (IL) / \emph{REWARDS (RL)} & EVALUATION & DATA \\
\hline

Bewley \emph{et al.} & transfers  & single frontal camera & steering & 1) image reconstruction loss 2) cyclic reconstruction loss & OPEN:\quad MAE, Balanced-MAE &  60K frames\\
 2019 \cite{bewley2019learning}& simulated to real & & &  3) control loss 4) cyclic control loss 5) adversarial loss & CLOSED: distance to intervention & town/rural \\
 
& (custom simulator) & & &  6) perceptual loss 7) latent reconstruction loss & & clear/rain/overcast \\

\hline

Codevilla \emph{et al.} & CARLA &	single frontal camera  & steering, acceleration &	MAE	&  & Train: 100h , Test: 80h \\
2019 \cite{codevilla2019exploring}& & ego-speed, navigation command & AUXILIARY: speed from vision & & CLOSED: CARLA benchmark\footnotemark  &  two towns, diverse weather \\

& & & & & NoCrash benchmark \footnotemark  &  \& traffic density\\

\hline
\setcounter{footnote}{0}

Chen \emph{et al.} & CARLA & single frontal camera & waypoints & MAE loss on trajectories & CLOSED: CARLA benchmark\footnotemark & Train: 154K frames = 4h \\
2019 \cite{chen2019learning} &  &  &(in the camera reference frame) & (comparing with privileged agent) & NoCrash benchmark\footnotemark & two towns, diverse weather\\

\hline
Hawke \emph{et al.} & real & 1 or 3 cameras & steering: value and slope & future-discounted MSE on  & OPEN: Balanced-MAE for model selection & Train: 30h, Test: 26 routes \\

2019  \cite{hawke2019urban} & & 2 timesteps (only for flow) & speed: value and slope & predicted values \& slopes  &  CLOSED: success \% of turns, of stopping behind pace car  & one city, 6 months  \\

&  & navigation command & & vs. observed future values &  collision rate, traffic violation rate & different times of day \\

& & & & & Meters per intervention (overall, in line follow, in pace car following) & \\

\hline
Hecker \emph{et al.}	& real	& single frontal camera & steering, speed	& MAE on speed and steering	& OPEN: MAE & Train: 60h, Test: 10h \\

2019 \cite{hecker2019learning}& & TomTom screen & & MAE on second derivative of speed and steering over time &  & city+countryside \\

& &  features from HD maps & & adversarial (log loss) on humanness of command sequence & & \\

\hline
Kendall	\emph{et al.} & real  & single frontal camera & steering, speed & Reward: distance travelled without driver taking over & CLOSED:	meters per disengagement &
250m rural road  \\

2019 \cite{kendall2019learning} & & & &  & & \\

\hline
\setcounter{footnote}{0}

Xiao \emph{et al.} & CARLA	& single frontal camera & steering, throttle, brake	& MAE	& CLOSED: CARLA benchmark\footnotemark &	Train: 25h \\
2019 \cite{xiao2019multimodal} &  & depth image (true or estimated) & &  & & two towns, diverse weather \\

\hline
Zhou \emph{et al.} & GTA V &  single or multiple cameras + & steering & MAE for steering & CLOSED: \% of successful episodes, & 	Train: 100K frames =  3.5 h \\

2019 \cite{zhou2019does}& & true or predicted: & & If predicting: MAE for depth, normals, flow, albedo & \% success weighted by track length & \\
& & semantic \& instance segmentation, monocular  & & CE for segmentation and boundary prediction &  & urban + off-road trail \\
& & depth, surface normals, optical flow, albedo & & & & \\

\hline
Zeng \emph{et al.} & real & single frontal camera & space-time cost volume,  & planning loss (cost maps) & OPEN: At different time horizons: &  1.4M frames, 6500 scenarios\\

2019  \cite{zeng2019end} & & LiDAR, ego-speed & AUXILIARY: ego-speed, 3D object &  perception loss (object detection) & MAE and MSE loss of trajectory  & multiple cities \\

& & & locations and future trajectories &  & collision \& traffic violation rate & \\
\hline

Amini \emph{et al.} & real	& 3 cameras, unrouted map, & 1) unrouted: weight, mean, variance  & unrouted: negative log-likelihood of human steering, & OPEN: z-score of human steering  &  Train: 25 km test: separate 1km \\

2018 \cite{amini2019variational} & &(optional) routed map & of the 3 gaussian models (GMs) & routed: steering MSE loss, & according to the GMM & suburban with turns, intersections, \\

& & &2) routed: deterministic steering control & MAE penalty on norm of the Gaussian  & &roundabouts,  dynamic obstacles \\ 
& & & & mixture model weights vector, & & \\
& & & & quadratic penalty on the log of variance of GMs & &  \\

\hline
Bansal \emph{et al.} & real, & 7 top down semantic maps  & waypoints, headings, speeds, self position & 1) waypoint (CE)  2) agent box loss (CE)  & OPEN: MSE on waypoints & 26M examples
= 60days \\

2018 \cite{bansal2018chauffeurnet}&  Test: real+sim & (Roadmap, Traffic Lights, Speed Limit,  &  ,  & 3) direction (MAE) 4) p\_subpixel loss (MAE) & CLOSED: success \% at stop-signs, traffic lights, lane following,   & no information on diversity \\ 

 & &  Route,  Current Agent Box, &  AUXILLIARY: road mask, perception boxes  & 5) speed (MAE) 6) collision loss 7)on road loss & navigating around a parked car, &\\

 & &  Dynamic Boxes,
Past Agent Poses) & & 8)geometry loss AUXILIARY: objects loss \& road loss & recovering from perturbations, slowing down behind a slow car & \\
\hline
\setcounter{footnote}{0}

Codevilla \emph{et al.}  &	CARLA + & single frontal camera & steering & MSE  &  CLOSED CARLA: success rate, km per infraction  & SIMULATION: train 2h,  two towns, diverse weather\\
2018 \cite{codevilla2018end} & real toy truck & ego-speed, navigation command &  acceleration &  & CLOSED REAL: \% missed turns, \# interventions, time spent & REAL: Train 2h, not diverse  \\
\hline

Hecker \emph{et al.}&	real & 4 cameras  & steering, speed & MSE & OPEN: MSE  & 60h \quad multiple cities, conditions \\
2018 \cite{hecker2018end} & & route: GPS coordinates or TomTom map & & & & \\
\hline
\setcounter{footnote}{0}

Liang \emph{et al.} & CARLA & single frontal camera & steering, acceleraton, brake & Trained in two phases,  & CLOSED: CARLA benchmark\footnotemark & IL: 14h + RL: 12h\\
2018 \cite{liang2018cirl}& & navigation command &  & In IL phase: \quad \quad  MSE & & two towns, diverse weather \\
 & & & & In RL phase: speed(+), abnormal steer angle (-) & &\\
  & & & & collisions (-), overlap with sidewalk or other lane (-) & &\\

\hline
Müller \emph{et al.} & transfers &  &  &  &  & Train: 28 h (in clear daytime weather) \\
2018 \cite{muller2018driving} & simulated to real & single frontal camera&  two waypoints (fixed distance, predicted angle)& MSE &CLOSED: \% of sucessful episodes  & Test: 2x25 trials \\
 & (simulation: CARLA) & navigation command & & &in real: time spent, missed turns, infractions & in cloudy after rain weather\\
 & (real: toy trucks) & & & & & two towns\\

 & & & & & & real: diverse situations, weather diversity unclear \\

\hline
\setcounter{footnote}{0}

Sauer \emph{et al.} & CARLA & single frontal camera & 6 affordances: 
& 3 x CE & CLOSED: CARLA benchmark\footnotemark
& no information on amount \\
2018 \cite{sauer2018conditional}& & & Hazard stop (boolean), Red Traffic Light (boolean),
& 3 x MAE & mean distance (km) between various types of infractions
& two towns, diverse weather\\

 & & & Speed Sign [categorical],  Relative angle [rad] & & in SI: distance to centerline, jerk &\\
 
 & & & Distances to vehicle [m] and centerline [m] & & &\\
\hline
Sobh \emph{et al.} & CARLA & single frontal camera as RGB & steering, throttle & MSE & CLOSED: time spent off-road  & Train: 136K samples  Test:20 min \\
2018 \cite{sobh2018end} & & or as segmentation & & & time spent on lane markings &\\
 & & LiDAR in BEV or PGM & & & number of crashes & weather not diverse (not specified) \\
 & & navigation command & & & &\\
\hline
\setcounter{footnote}{0}

Dosovitsky \emph{et al.} & CARLA & single frontal camera & for IL model: specified as "action" & not specified & CLOSED: CARLA benchmark\footnotemark & IL model:  14h of driving data
\\
2017 \cite{dosovitskiy2017carla}& &navigation command & RL model: no information &  & distance between infractions & RL model: 12 days of driving\\
\hline
Bojarski \emph{et al.} & real & single frontal camera & steering & MSE & CLOSED: autonomy (\%of driving time  & Train: length not specified \\

2016 \cite{bojarski2016end}& & & & & when car was controlled by the model, not safety driver) & day, night, multiple towns, conditions \\
&  & & & &  & test: 3h = 100km\\
\hline

\end{tabular}
\end{tiny}
\end{table}

\setcounter{footnote}{1}
\footnotetext{CARLA benchmark. The model is tested on routes of varying difficulty, with and without dynamic objects, while the percentage of successful routes and kilometers per infraction are measured. There are two towns and 6 different weather conditions. One of the towns and two of the weather conditions are novel, i.e. excluded from training set. }
\setcounter{footnote}{2}

\footnotetext{NoCrash benchmark. The model is tested on routes in 3 different traffic densities, 6 different weather conditions and two different towns. Percentage of successful episodes is counted. One of the towns and two of the weather conditions are novel, i.e. excluded from training set. }
 
\end{landscape}

\begin{landscape}
\section{Benchmark results for CARLA simulator}
Standardized benchmarking of end-to-end driving models is available only in the CARLA simulator. In Tables \ref{tab:CARLA_bench} and \ref{tab:NoCrash} we summarize the generalization ability for the two benchmarks available in CARLA: original and NoCrash. We have omitted the benchmark results in the training town, Town 1, and only report performance in Town 2.

\begin{table}[h!]
    \centering
    \caption{CARLA benchmark results in Town 2 (not seen during training), in weather conditions either seen or not seen during training and in the absence or presence of other dynamic objects (pedestrians, cars). Unreported values are marked with - .}
    \begin{tabular}{|c|c|c||c|c||c|c||c|c|}
    \hline
         & \multicolumn{4}{c||}{Town I} & \multicolumn{4}{c|}{ Town II}\\
        ARTICLE & \multicolumn{2}{c||}{Training weather} & \multicolumn{2}{c|}{New weather} & \multicolumn{2}{c||}{Training weather} & \multicolumn{2}{c|}{New weather}\\

         & Empty  &  Dyn. objects & Empty & dyn. objects  & Empty  &  Dyn. objects & Empty & dyn. objects \\ \hline
        Chen \emph{et al.} 2019 \cite{chen2019learning}& 100 \textsuperscript{a} & 100 \textsuperscript{a} & 100 \textsuperscript{a} & 100 \textsuperscript{a} & 100 & 99 &100 &100 \\\hline
        Xiao \emph{et al.} 2019 \cite{xiao2019multimodal} \textsuperscript{b} & 92 & 89 &90 & 90& 90 & 87 & 90 & 94\\\hline
        Codevilla \emph{et al.} 2019 \cite{codevilla2019exploring} & 95 & 92 & - & - & 95 & 92 & 92 & 90 \\\hline
        Sauer \emph{et al.} 2018 \cite{sauer2018conditional} & 92 & 93 & 90 & 92 & 70 &64 & 68 & 64\% \\\hline
        Liang \emph{et al.} 2018 \cite{liang2018cirl} & 93 & 82 & 86 & 80 & 53 & 41 & 68 & 62 \\\hline
        Dosovitsky \emph{et al.} 2017 \cite{dosovitskiy2017carla} & 86 & 83 & 84 & 82 & 40 & 38 & 44 & 42 \\\hline
        \end{tabular}\\
\vspace{0.5cm}   
\textsuperscript{a} As reported in the GitHub repository accompanying the article (https://github.com/dianchen96/LearningByCheating)

\quad \quad \quad \quad \textsuperscript{b} Xiao et al. \cite{xiao2019multimodal} is the only surveyed CARLA-based model that uses depth information. The reported results are obtained by a model that early-fuses a perfect\newline depth image (obtained directly from the simulator) with RGB image.

\label{tab:CARLA_bench}
\end{table}

\begin{table}[h!]
    \centering
    \caption{Results for different models in NoCrash benchmark in CARLA simulator (version $<$0.9.5), as reported in \cite{chen2019learning} and \cite{codevilla2019exploring}. Only the performance in the Town 2 (town not observed during training) is reported here, as it offers a fairer measure of generalization ability. }
    \begin{tabular}{|c|c|c|c||c|c|c|}
    \hline
         & \multicolumn{3}{c||}{Training weather} &\multicolumn{3}{c|}{New weather}\\
        ARTICLE & No traffic & Regular traffic & Dense traffic & No traffic & Regular traffic & Dense traffic \\ \hline
        Chen \emph{et al.} 2019 \cite{chen2019learning}& 100 & 96&89 & 100\% & 94\% &85\% \\\hline

        Codevilla \emph{et al.} 2019 \cite{codevilla2019exploring} & 51 & 44 & 38  & 90\% & 87\% & 67\% \\\hline 
        
        Sauer \emph{et al.} 2018 \cite{sauer2018conditional} & 36 & 26 & 9 & 25\% &14\% & 10\%  \\\hline
        
        Codevilla \emph{et al.} 2018 \cite{codevilla2018end} & 48 &27 &10 & 24\% & 13\% & 2\%\\\hline
    \end{tabular}
    \label{tab:NoCrash}
\end{table}
\end{landscape}

\begin{landscape}

\section{Datasets}
\label{sec:Datasets}

\renewcommand*{\thefootnote}{\fnsymbol{footnote}}
\setcounter{footnote}{0}
Studies presented in the Appendix A have either i) access to an experimental platform allowing to collect their own real-world dataset or ii) perform closed-loop evaluation. The collected datasets are rarely made public. This creates a problem for researchers who do not have a possibility to produce their own data. This appendix lists publicly available datasets which can be used by the wider research community. Therefore the connection between methods in Appendix A and the datasets here is negligible. 

\begin{table}[h]
\centering
\caption{List of useful datasets for training end-to-end self-driving models. Abbreviations introduced in this table: global navigation satellite system (GNSS).}

\begin{tabular}{|l|c|c|c|c|c|c|c|c|c|c|c|c|c|c|c|c|}
\hline
\multirow{2}{*}{Dataset} & \multicolumn{7}{c|}{Modalities} & \multicolumn{4}{c|}{Annotations} & \multicolumn{3}{c|}{Diversity} & \multirow{2}{*}{Size} & \multirow{2}{*}{License} \\
\cline{2-15}

& \rotatebox{90}{\parbox{2cm}{Cameras}}
& \rotatebox{90}{\parbox{2cm}{LiDAR}}
& \rotatebox{90}{\parbox{2cm}{GNSS}}
& \rotatebox{90}{\parbox{2cm}{Steering}}
& \rotatebox{90}{\parbox{2cm}{Speed, \\ Acceleration}}
& \rotatebox{90}{\parbox{2cm}{Navigational \\ command}}
& \rotatebox{90}{\parbox{2cm}{Route planner}}
& \rotatebox{90}{\parbox{2cm}{Semantic map}}
& \rotatebox{90}{\parbox{2cm}{Text annotations}}
& \rotatebox{90}{\parbox{2cm}{3D annotations}}
& \rotatebox{90}{\parbox{2cm}{2D annotations}}
& \rotatebox{90}{\parbox{2cm}{Weather}}
& \rotatebox{90}{\parbox{2cm}{Time of day}}
& \rotatebox{90}{\parbox{2cm}{Driving scene}}
& 
& \\
\hline

Udacity \cite{udacity_dataset:2017} & 3 & \checkmark & \checkmark & \checkmark & \checkmark & & & & & & & \checkmark & & & 5h & MIT \\
\hline
CARLA \cite{dosovitskiy2017carla} & 1 & & \checkmark & \checkmark & \checkmark & \checkmark & & & & & \checkmark & \checkmark & & & 12h & MIT \\
\hline
Drive360 \cite{hecker2018end} & 8 & & \checkmark & \checkmark & \checkmark & & \checkmark & \checkmark & & & & \checkmark & \checkmark & \checkmark & 55h & Academic \\
\hline
Comma.ai 2016 \cite{santana2016learning} & 1 & & \checkmark & \checkmark & \checkmark & & & & & & & \checkmark & \checkmark & & 7h 15min & CC BY-NC-SA 3.0 \\
\hline
Comma.ai 2019 \cite{schafer2018commute} & 1 & & \checkmark & \checkmark & \checkmark & & & & & & & & \checkmark & & 30h & MIT \\
\hline
DeepDrive \cite{yu2018bdd100k} & 1 & & \checkmark & & & & & & \checkmark & & \checkmark & \checkmark & \checkmark & \checkmark & 1100h & Berkley \\
\hline
DeepDrive-X \cite{kim2018textual} & 1 & & \checkmark & & & & & & \checkmark & & & \checkmark & \checkmark & \checkmark & 77h & Berkley \\
\hline
Oxford RobotCar \cite{maddern20171} & 4 & \checkmark & \checkmark & & & & & & & & & \checkmark & \checkmark & \checkmark & 214h & CC BY-NC-SA 4.0 \\
\hline
HDD \cite{ramanishka2018toward} & 3 & \checkmark & \checkmark & \checkmark & \checkmark & \checkmark & & & \checkmark & & & \checkmark & \checkmark & \checkmark & 104h & Academic \\
\hline
Brain4Cars \cite{jain2015car} & 1 & & \checkmark & & \checkmark & & & & \checkmark & & & & & \checkmark & 1180 miles & Academic \\
\hline
Li-Vi \cite{chen2018lidar, dbnet:2018} & 1 & \checkmark & \checkmark & \checkmark & \checkmark & & & & & & & & & \checkmark & 10h & Academic \\
\hline
DDD17 \cite{binas2017ddd17} & 1\footnotemark & & \checkmark & \checkmark & \checkmark & & & & & & & \checkmark & \checkmark & \checkmark & 12h & CC-BY-NC-SA-4.0 \\
\hline
A2D2 \cite{geyer2019a2d2} & 6 & \checkmark & \checkmark & \checkmark & \checkmark & & & & & \checkmark & \checkmark & & & & 390k frames & CC BY-ND 4.0 \\
\hline
nuScenes \cite{caesar2019nuscenes} & 6 & \checkmark & \checkmark & & & & & \checkmark & & \checkmark & \checkmark & \checkmark & \checkmark & \checkmark & 5.5h & Non-commercial \\
\hline
Waymo \cite{sun2019scalability} & 5 & \checkmark & \checkmark & \checkmark & \checkmark & & & & & \checkmark & \checkmark & & \checkmark & & 5.5h & Non-commercial \\
\hline
H3D \cite{patil2019h3d} & 3 & \checkmark & \checkmark & \checkmark & \checkmark & & & & & \checkmark & & & & & N/A & Academic \\
\hline
HAD \cite{kim2019grounding} & 3 & & \checkmark & \checkmark & \checkmark & \checkmark & & \checkmark & \checkmark & & & \checkmark & & & 30h & Academic \\
\hline
\end{tabular}
\label{tab:datasets}
\end{table}
\footnotetext{Event camera}
\end{landscape}


\section*{Acknowledgments}

The authors would like to thank Hannes Liik for fruitful discussions.

Ardi Tampuu, Maksym Semikin, Dmytro Fishman and Tambet Matiisen were funded by European Social Fund via Smart Specialization project with Bolt. Naveed Muhammad has been funded by European Social Fund via IT Academy programme.

\ifCLASSOPTIONcaptionsoff
  \newpage
\fi



%

\bibliographystyle{unsrt} 
\bibliography{references} 


%
\begin{IEEEbiography}[{\includegraphics[width=1in,height=1.25in,clip,keepaspectratio]{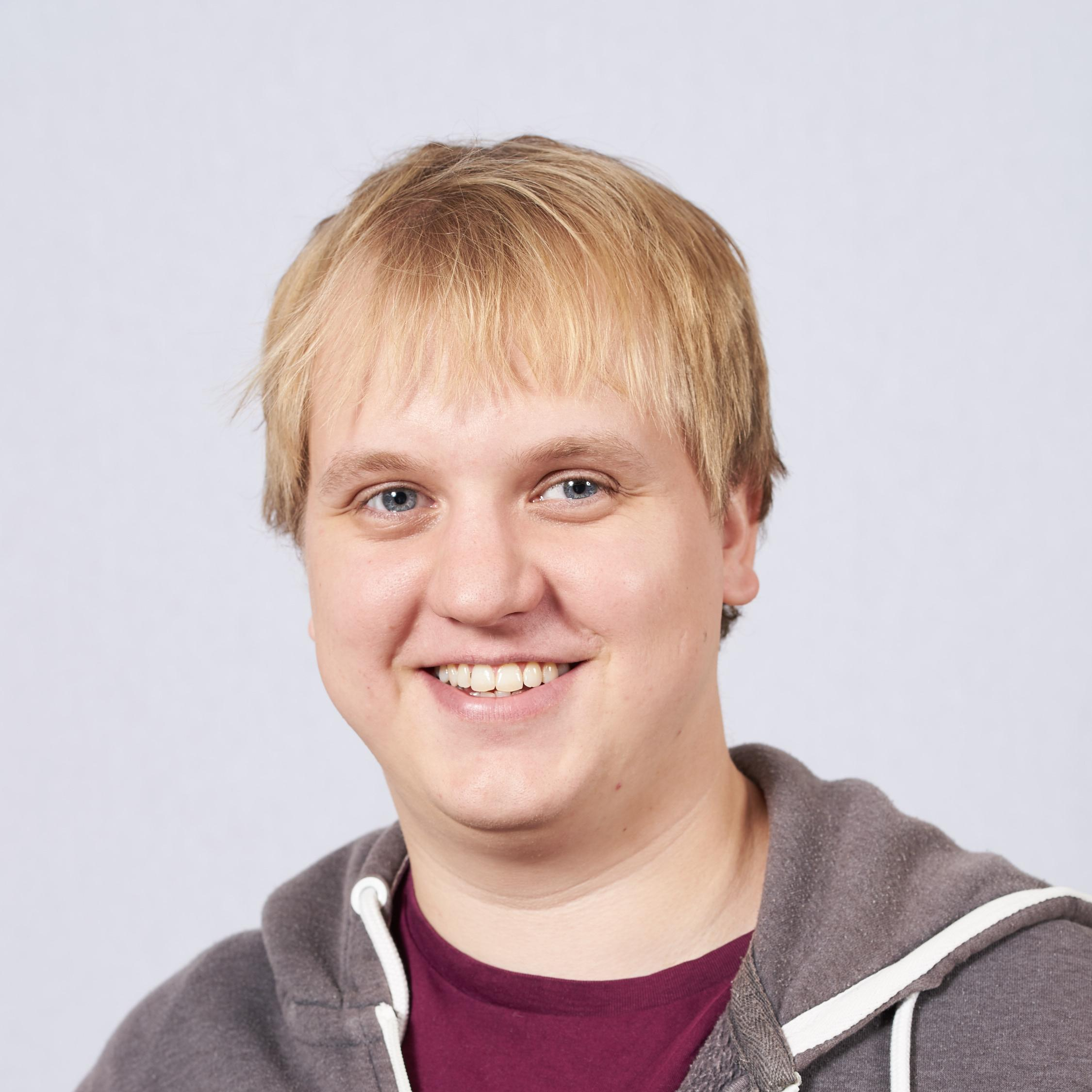}}]{Ardi Tampuu}
is working as a research software engineer and is leading the research effort on end-to-end driving in Autonomous Driving Lab, at the Institute of Computer Science, University of Tartu, Estonia. He earned his PhD from University of Tartu in 2020. Earlier, he earned his engineering and MSc degrees from INSA de Lyon, France. He has extensive experience in teaching and supervision of neural networks (NNs) courses and projects, in several application domains. 
\end{IEEEbiography}
\vspace{-1cm}
\begin{IEEEbiography}[{\includegraphics[width=1in,clip,keepaspectratio]{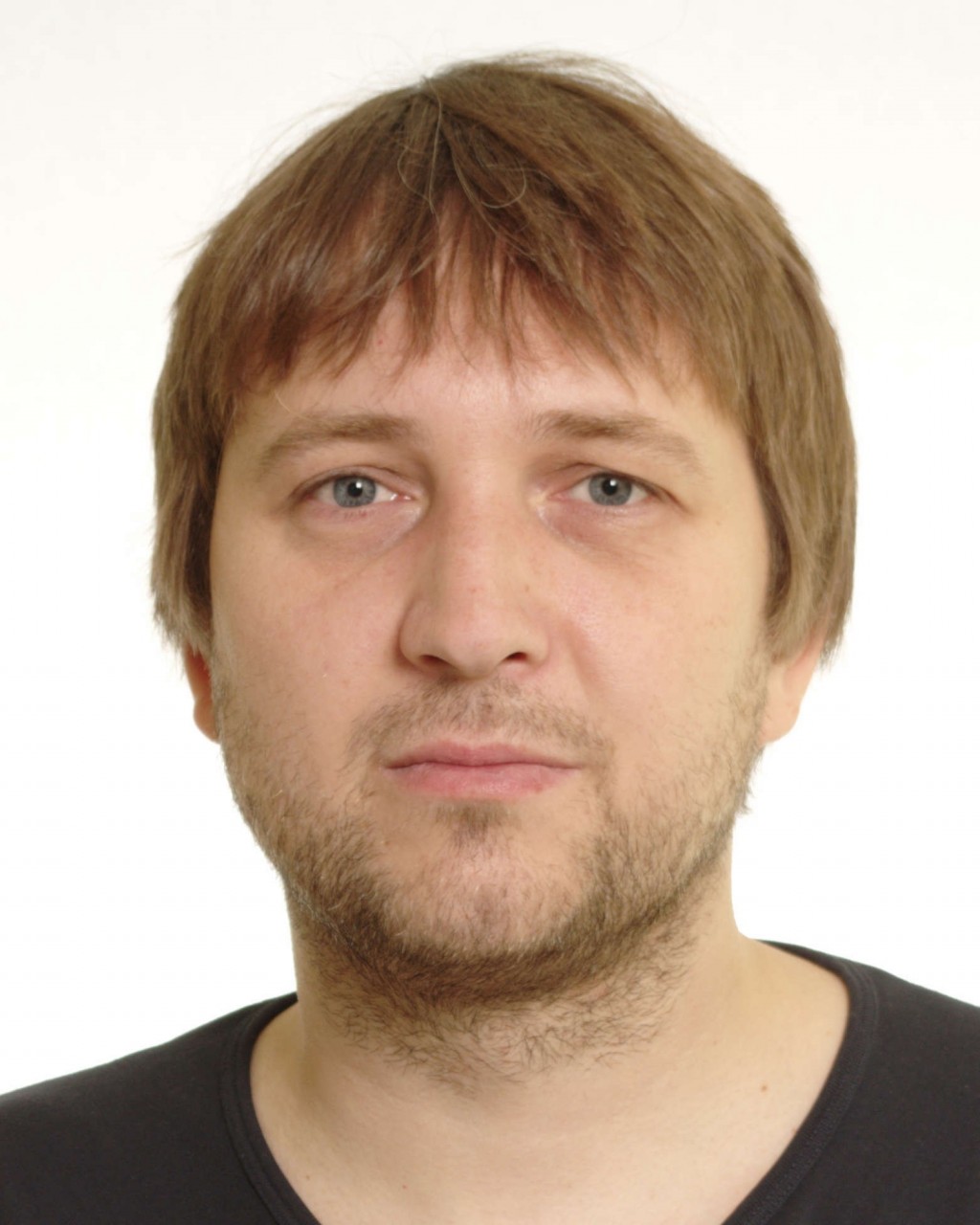}}]{Tambet Matiisen}
leads the tech. side of the Autonomous Driving Lab at the Institute of Computer Science, University of Tartu, Estonia. At the lab, he has been involved in a number of projects with industrial partners including Milrem Robotics and Bolt, and is also finishing his PhD in deep reinforcement learning. In the past he has worked on automated curriculum learning, at OpenAI. His research interests include end-to-end autonomous driving, curriculum learning and intrinsic motivation. 
\end{IEEEbiography}
\vspace{-1cm}
\begin{IEEEbiography}[{\includegraphics[width=1in,clip,keepaspectratio]{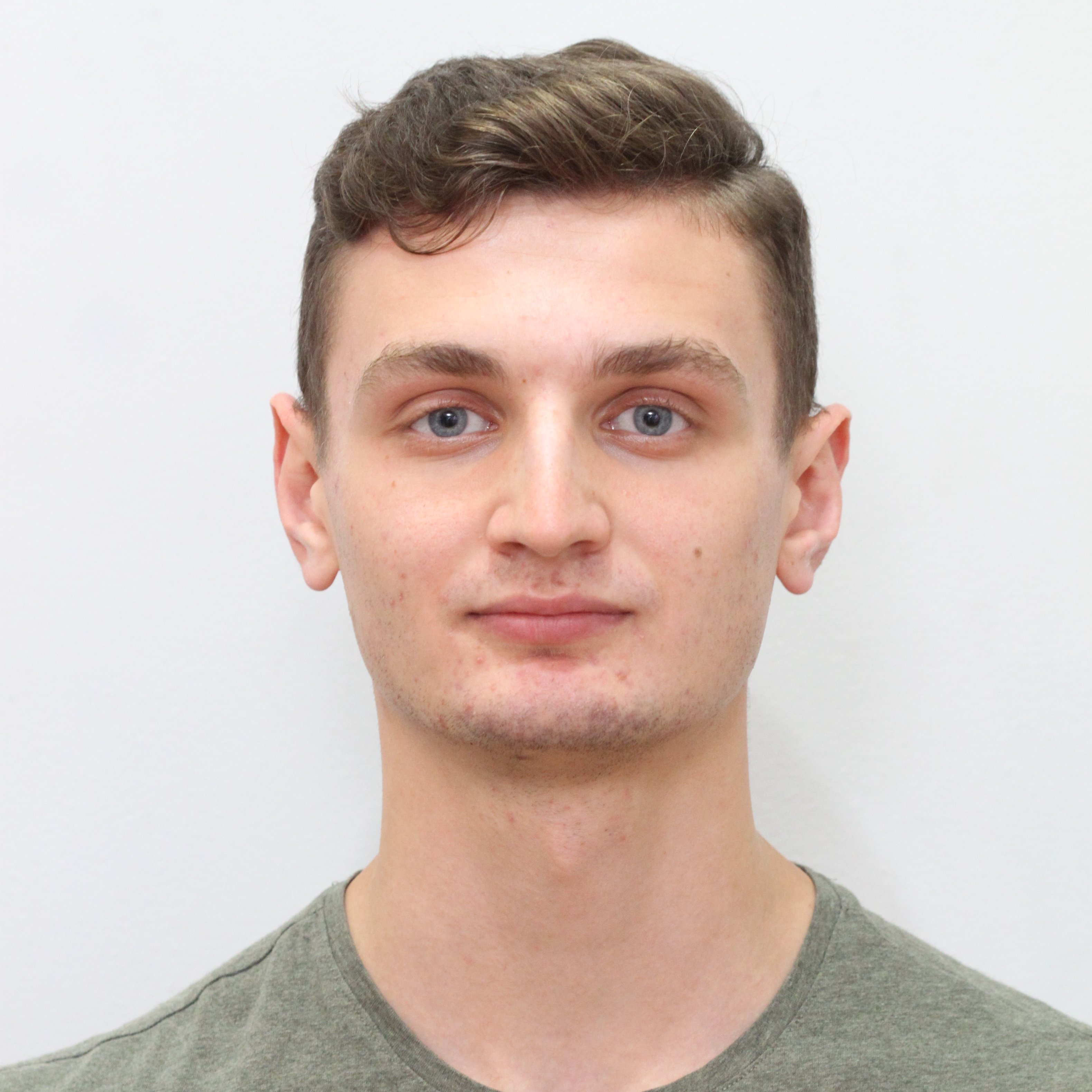}}]{Maksym Semikin}
has worked as a software engineer in multiple domains as well as a Data Scientist focusing on content personalization in E-commerce. He earned his MSc in Computer Science, with specialization in Data Science, from University of Tartu in 2019 and was a member of the University of Tartu Autonomous Driving Lab. His research interests include machine learning and its applications.
\end{IEEEbiography}
\vspace{-1cm}
\begin{IEEEbiography}[{\includegraphics[width=1in,clip,keepaspectratio]{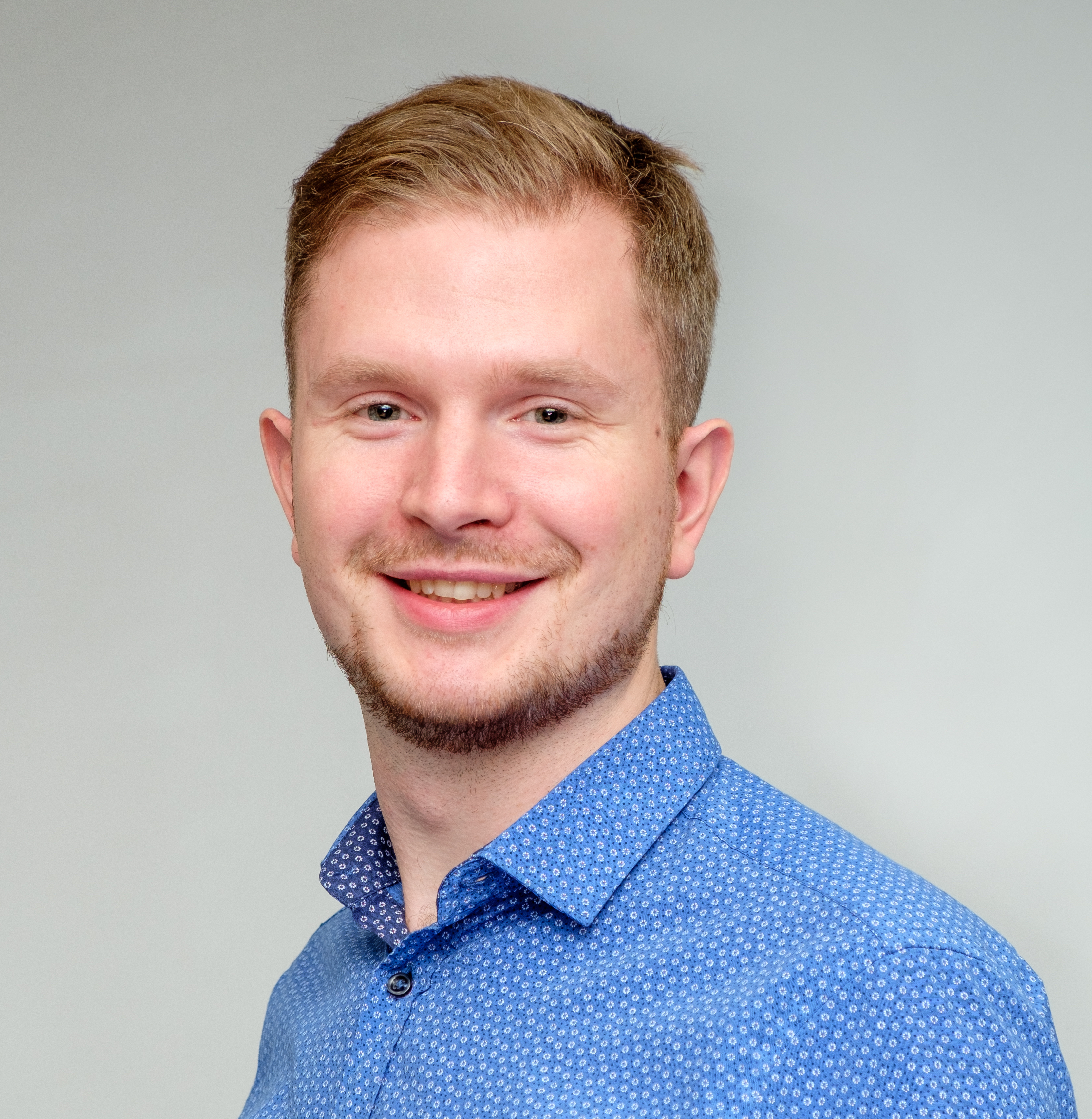}}]{Dmytro Fishman}
is working as junior lecturer of data science at the Institute of Computer Science, University of Tartu, Estonia. At the university’s Autonomous Driving Lab, he heads the mapping and localization team. In addition to working in collaboration with Bolt on autonomous driving, he has collaborated with industrial partners such as PerkinElmer. He has a keen interest in data-science education and is a certified instructor at Data and Software Carpentry organisations that organise and carry out data science training for researchers around the world. His research interests include machine learning, deep learning and artificial intelligence.
\end{IEEEbiography}
\vspace{-1cm}
\begin{IEEEbiography}[{\includegraphics[width=1in,clip,keepaspectratio]{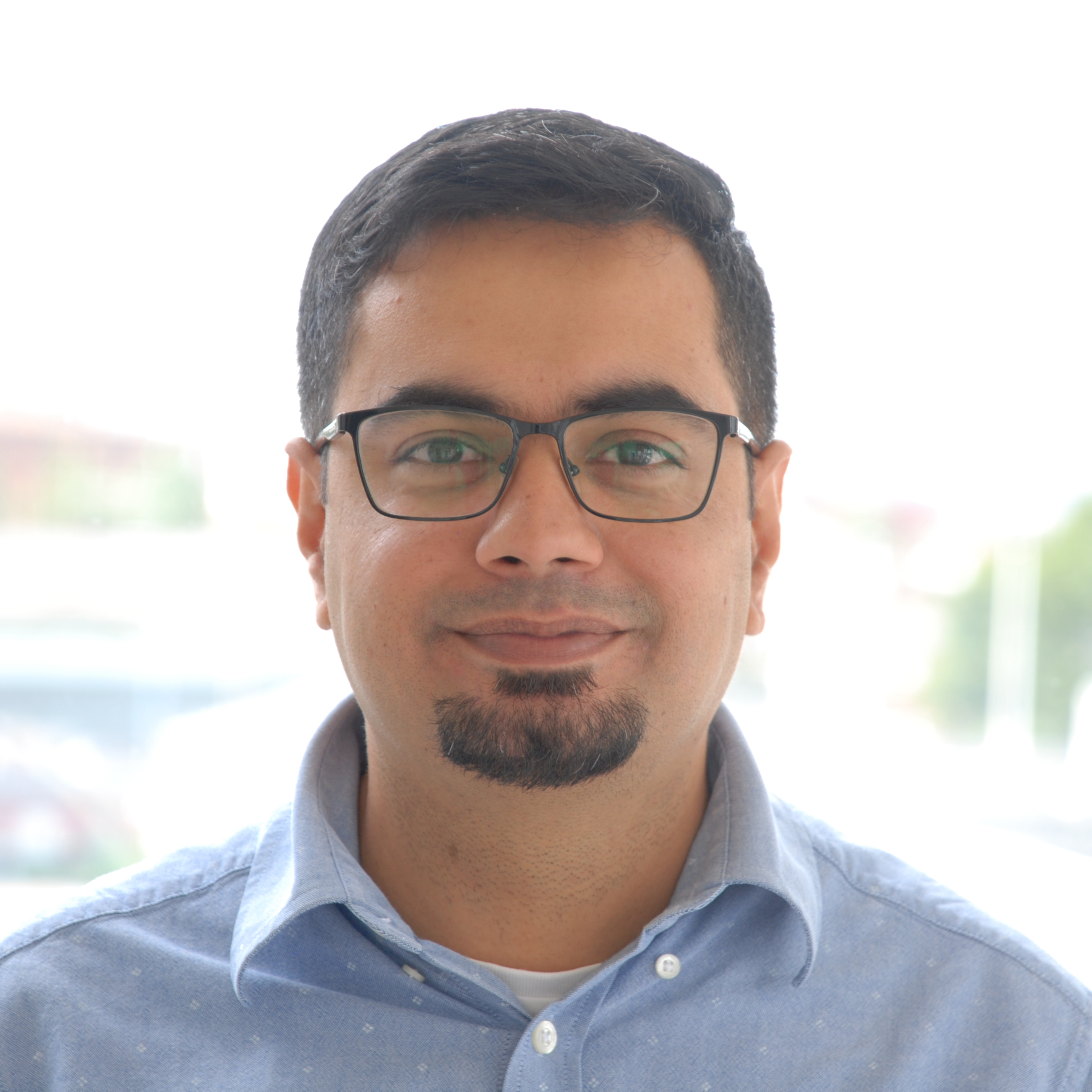}}]{Naveed Muhammad}
is working as assistant professor of autonomous driving at the University of Tartu, Estonia, where he is also a part of the Autonomous Driving Lab. He earned his PhD in robotics from INSA de Toulouse (research stay at LAAS-CNRS), France, in 2012. He has had postdoctoral stays at Tallinn University of Technology, Estonia, and Halmstad University, Sweden, and has taught at National University of Sciences and Technology, Pakistan and Asian Institute of Technology, Thailand. His research interests include autonomous driving, perception, and behaviour modeling. 
\end{IEEEbiography}




\end{document}